\documentclass[10pt,twocolumn,letterpaper]{article}

\usepackage[pagenumbers]{cvpr}

\usepackage[dvipsnames]{xcolor}

\definecolor{cvprblue}{rgb}{0.21,0.49,0.74}
\usepackage[unicode=true,pagebackref,breaklinks,colorlinks,citecolor=cvprblue]{hyperref}
\usepackage{natbib}
\setcitestyle{square, comma, numbers,sort&compress, super}
\let\turc\c
\usepackage{etoolbox}
\robustify\turc

\renewcommand{\c}{\ensuremath{c^\pi}}

\usepackage{graphicx}
\usepackage{amsmath}
\usepackage{amssymb}
\usepackage{booktabs}
\usepackage{bm}
\usepackage{adjustbox}
\usepackage{mathtools}
\usepackage{amsfonts}
\usepackage{array}
\usepackage{tikz}
\usepackage{ctable}
\usepackage{float}
\usepackage{enumitem}
\usepackage{dsfont}
\usepackage{xcolor}         
\usepackage{dashrule}       
\usepackage{makecell}       
\usepackage{colortbl}       
\usepackage{times}
\usepackage{multirow}
\usepackage{subcaption}

\allowdisplaybreaks

\definecolor{blueblack}{RGB}{0, 108, 173}
\definecolor{taborange}{RGB}{235, 127, 14}
\definecolor{tabgreen}{RGB}{30, 160, 30}
\definecolor{tabpurple}{RGB}{128, 103, 189}

\definecolor{sol_light_blue}{RGB}{38, 139, 210}
\definecolor{sol_blue}{RGB}{38, 139, 210}
\definecolor{nord_blue}{RGB}{38, 139, 210}
\definecolor{sol_green}{RGB}{163, 190, 140}
\definecolor{sol_red}{RGB}{220, 50, 47}
\definecolor{nord_red}{RGB}{250, 190, 192}
\definecolor{nord_green}{RGB}{163, 190, 140}

\definecolor{beer_orange}{RGB}{242, 142, 28}

\definecolor{nordblack}{RGB}{46, 52, 64}
\definecolor{nordred}{RGB}{191, 97, 106}
\definecolor{magenta}{RGB}{215, 10, 185}
\definecolor{nordgreen}{RGB}{143, 170, 120}
\definecolor{nordblue}{RGB}{94, 129, 172}
\definecolor{nordpurple}{RGB}{180, 142, 160}

\usepackage[capitalize]{cleveref}
\crefname{section}{Sec.}{Secs.}
\Crefname{section}{Section}{Sections}
\Crefname{table}{Table}{Tables}
\crefname{table}{Tab.}{Tabs.}

\newcommand{\A}{\mathbf{A}}

\renewcommand{\c}{\mathbf{c}}
\newcommand{\x}{\mathbf{x}}
\newcommand{\y}{\mathbf{y}}

\newcommand{\e}{\boldsymbol{\epsilon}}

\newcommand{\p}{\mathbf{p}}

\renewcommand{\b}{\mathbf{b}}

\newcommand{\X}{\mathbf{X}}

\newcommand{\W}{\mathbf{W}}

\renewcommand{\e}{\mathbf{e}}

\newtheorem{lem}{Lemma}

\title{Robust Point Cloud Processing through Positional Embedding}
\author{Jianqiao Zheng$^{1}$\thanks{Corresponding email: jianqiao.zheng@adelaide.edu.au. 
Project page at \href{https://osiriszjq.github.io/RobustPPE}{https://osiriszjq.github.io/RobustPPE}.}
\and
Xueqian Li$^{1}$
\and
Sameera Ramasinghe$^{2}$
\and
Simon Lucey$^{1}$
\and
\\
\begin{tabular}[h]{cc}
	$^{1}$The University of Adelaide \quad\quad $^{2}$Amazon \\
\end{tabular}
}

\begin{document}
\maketitle

\begin{abstract}
   End-to-end trained per-point embeddings are an essential ingredient of any state-of-the-art 3D point cloud processing such as detection or alignment. 
   Methods like PointNet~\cite{qi2017pointnet}, or the more recent point cloud transformer~\cite{guo2021pct}---and its variants---all employ learned per-point embeddings. 
   Despite impressive performance, such approaches are sensitive to out-of-distribution (OOD) noise and outliers. 
   In this paper, we explore the role of an analytical per-point embedding based on the criterion of bandwidth. 
   The concept of bandwidth enables us to draw connections with an alternate per-point embedding---positional embedding, particularly random Fourier features. 
   We present compelling robust results across downstream tasks such as point cloud classification and registration with several categories of OOD noise.
\end{abstract}

\vspace{-0.3cm}
\section{Introduction}
\label{sec:intro}

\begin{figure}[t]
    \centering
    {\includegraphics[width=0.95\linewidth]{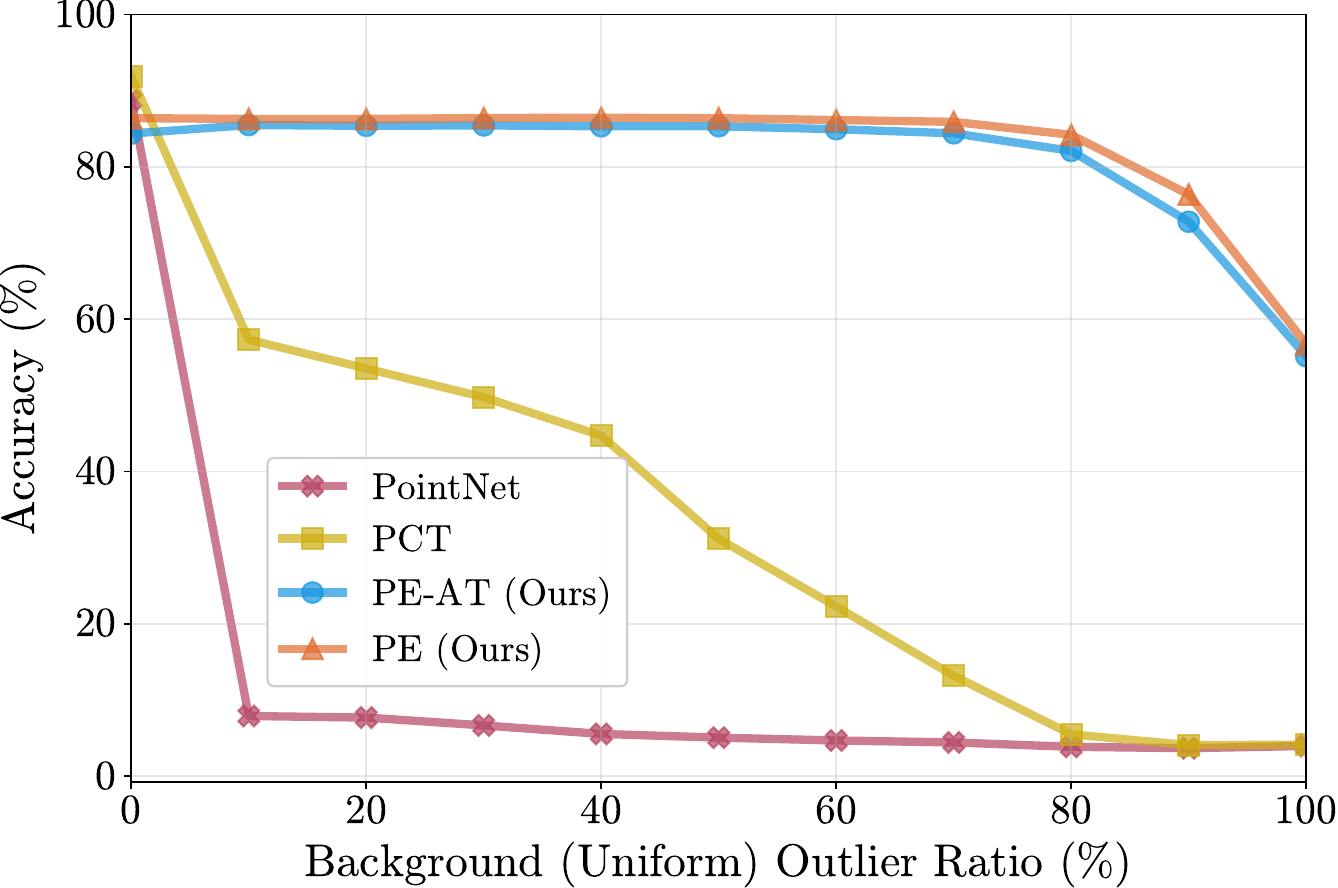} }%
    \vspace{-0.2cm}
    \caption{The robustness of positional embedding (PE)-based methods (PE-AT and PE) in comparison to end-to-end trained methods (PointNet and PCT). 
    We show the classification accuracy on ModelNet40-C~\cite{sun2022benchmarking} versus the number of outliers (ratio to the number of point clouds) uniformly sampled from $[-1,1]^3$.
    We argue in this paper that PE, especially random Fourier features (RFF) embodies many of the properties one needs in a robust embedding without the need for learning, making it more robust than end-to-end trained per-point embeddings (PPEs). 
    }%
    \label{fig:gaussian_noise}%
    \vspace{-0.5cm}
\end{figure}

PointNet~\cite{qi2017pointnet} has been a critical innovation in the area of deep geometry, providing much of the backbone to innovations in object classification, segmentation, detection~\cite{qi2018frustum}, and alignment~\cite{aoki2019pointnetlk, li2021pointnetlk} on 3D point clouds. 
Many variants~\cite{qi2017pointnet++, wu2019pointconv, wang2019dynamic, thomas2019kpconv, zhao2021point, guo2021pct} have followed, most notably, point cloud transformer (PCT)~\cite{guo2021pct}~\footnote{We acknowledge that there is concurrent similar work such as point transformer~\cite{zhao2021point}. 
In this paper, we will focus on PCT~\cite{guo2021pct}.} that combines the strengths of PointNet and transformer networks into a state-of-the-art framework for 3D point cloud processing. 
All these variants have a common element---a learned---typically deep ReLU network---per-point embedding (PPE).

In this paper, we explore the hypothesis of whether a more robust performance can be afforded by frameworks such as PointNet or PCT if we replace the learned PPEs with analytical embeddings. 
This hypothesis has two motivations. 
First, we hypothesize that the earliest modules of modern point cloud networks---specifically the PPE---such as those found in PointNet and PCT, should be most generic in task and domain. 
As a result, learning such PPEs in an end-to-end fashion solely from data may make the entire pipeline overly sensitive to changes in the point cloud statistics~\cite{sun2022benchmarking}. 
Second, the success of positional embedding (PE) in machine learning frameworks such as transformer networks~\cite{vaswani2017attention} and implicit neural functions~\cite{mildenhall2021nerf} offers an alternate surrogate for end-to-end PPEs. 
PE~\cite{zheng2022trading} shares similar characteristics to learned PPEs in that they attempt to transform a scalar or even vector position into a multi-dimensional embedding. 
They differ fundamentally, however, in that the entire embedding is analytical requiring only a single tunable hyper-parameter---bandwidth~\cite{zheng2022trading}.

The success of PointNet/PCT is fundamentally wedded to the data it is trained upon. 
When the distribution of the test environment differs substantially from the training dataset, catastrophic performance can occur (see~\cref{fig:gaussian_noise} red and yellow curve). 
PointNet/PCT involves two components:
(i) a per-point high-dimensional embedding (PPE), which refers to all components before the pooling function, and 
(ii) a pooling operation that aggregates a global feature vector.
The theoretical role of (ii) is well understood as a mechanism for providing invariance to permutation~\cite{soelch2019deep}. 
In contrast, the role of (i) is largely shrouded in mystery since most of the community treats it as an opaque black box that must be learned. 
In this paper, we attempt to theoretically characterize the idealized properties of (i). 
In particular, we argue that such PPEs should be band-limited and smooth. 
Further, we argue that learned deep ReLU embeddings---commonly used in most implementations of PointNet/PCT---are inherently flawed in this regard. 
To make up for this shortcoming they are unnecessarily reliant on aggressive aggregation operators like max-pool to ensure good performance.

The PPEs used in techniques such as PointNet and PCT typically take the form of ReLU-MLPs. 
The weights of these ReLU-MLPs are learned end-to-end for the task at hand (\eg, alignment, detection,~\etc).
Although the data-dependent training implicitly achieves appropriate bandwidth, it inherently produces in-distribution bias. 
Inspired by positional embeddings (PE)---recently made popular in transformer networks~\cite{vaswani2017attention} and implicit neural functions such as NeRF~\cite{mildenhall2021nerf}---we explore the use of sinusoid-based PEs as a surrogate for learned PPEs in PointNet/PCT. 
Unlike deep ReLU, the per-point high dimensional embedding provided by PEs does \textit{not} need to be trained on any data---simply tuning the variance of the randomly initialized weights yields effective performance.
Our claims are further supported by strong empirical evidence by showcasing the performance of PE-based methods to the ReLU-MLPs on the object classification task (\cref{fig:gaussian_noise}).
Comparable performance is achieved when testing these methods in the in-distribution data domain (\ie, background outlier ratio is 0\%).
However, the real advantage of PEs becomes apparent when applied to more practical out-of-distribution (OOD) circumstances. 
As the mismatch between the train and test distributions increases, PE-based embeddings significantly outperform current learning methods by a large margin.

We conduct substantial experiments across popular benchmarks---ModelNet40~\cite{wu20153d} and ModelNet40-C~\cite{sun2022benchmarking}---and we show that one can achieve remarkably robust classification performance in the presence of various unseen data corruptions, such as noise and outliers.
Our paper makes the following contributions:
\begin{enumerate}
    \item We theoretically show why RFF-based PEs are excellent candidates for use as PPEs within PointNet and PCT. 
    In particular, we connect the variance of the weights to the bandwidth of the embeddings and spatial locality---speculating upon their utility in the presence of noise and outliers. 
    \item Strong empirical evidence that demonstrates that PE-based embeddings significantly outperform current end-to-end trained models when evaluated in a variety of OOD data corruption conditions on both classification and alignment tasks. 
\end{enumerate}

\section{Related works}
\label{sec:related_work}
\noindent\textbf{Learned PPE.}\:\: Approaches such as PointNet~\cite{qi2017pointnet} and DeepSets~\cite{zaheer2017deep} opened a new direction of directly processing unsorted 3D point cloud sets. 
Usually, this architecture uses MLPs to process the points, followed by a permutation invariant aggregation function. 
PointNet++~\cite{qi2017pointnet++} enhances its ability to capture local information and get impressive results.
Recently, point cloud transformer (PCT)~\cite{guo2021pct} combined the attention mechanism into per-point encoder. 
However, all these models need to be trained end-to-end. In a recent work~\cite{sanghi2020powerful} it was demonstrated that good performance could be achieved with a randomly initialized deep ReLU PPE within PointNet.

\noindent\textbf{Positional Encoding.}\:\:Transformers~\cite{vaswani2017attention} heavily utilized PE so as to bake positional information into the attention mechanism. 
The popular Neural radiance fields (NeRF)~\cite{mildenhall2021nerf} framework also use PE for its coordinate encodings. 
In fact, 2D coordinate networks already widely used positional encodings in tasks such as image reconstruction~\cite{nguyen2015deep, stanley2007compositional}.
Of particular note is the work of Mildenhall~\etal~\cite{mildenhall2020nerf} and Zhong~\etal~\cite{zhong2019reconstructing} who found empirically that using sinusoidal positional encodings allow networks to converge faster, and learn high-frequency signal content better. 
The most popular PE method in literature is random Fourier features (RFF)~\cite{tancik2020fourier} -- the authors motivate their mechanism of positional encoding through NTK theory~\cite{arora2019fine, bietti2019inductive, du2018gradient, jacot2018neural,lee2019wide}. 
Subsequent works have offered expanded theoretical motivations for PE~\cite{zheng2021rethinking} and~\cite{zheng2022trading} which use rank and smoothness to evaluate the utility of a positional encoding scheme.

\begin{figure*}[t]
\centering
\includegraphics[width=1.0\textwidth]{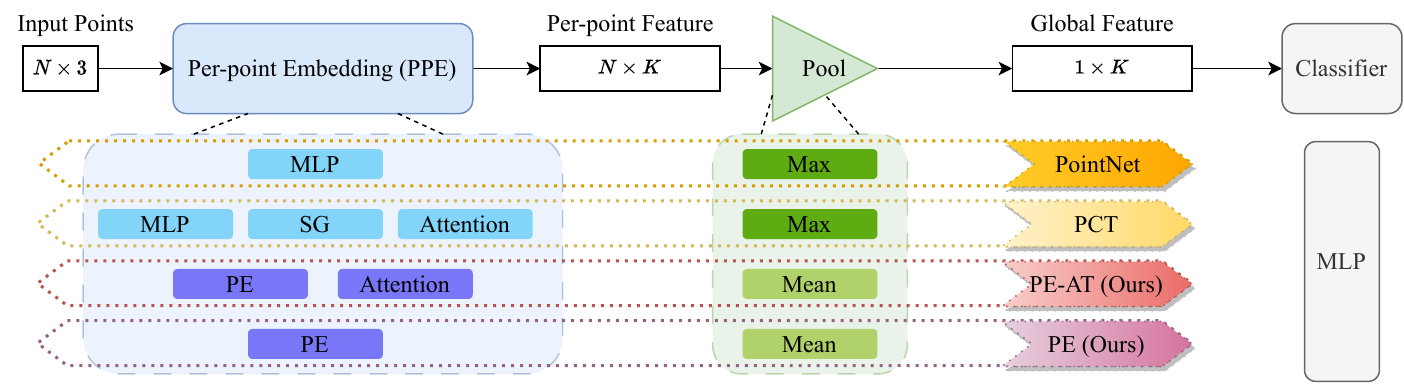}
\caption{Illustration of different point cloud processing architectures. 
SG: sampling and grouping. 
PE: positional encoding. 
In per-point embedding (PPE), light blue PPEs are trained end-to-end while purple ones are randomly initialized.
Both PointNet-based and PCT-based methods process points with a per-point encoder first and then use a pooling (usually max-pool) to get a feature vector for the point cloud. 
We found that all the pre-processing stages in PCT could be replaced by a simple PE (our PE-AT model) while maintaining good robustness to OOD noise. 
Furthermore, we can use PE as a non-learned PPE (our model) and achieves comparable performance. 
A pooling operation is used to get the global feature vector.
And all these models can be used in various downstream tasks, such as classification.}
\vspace{-0.2cm}
\label{fig:method}
\end{figure*}

\noindent\textbf{Pooling Function.}\:\: In the PointNet architecture, the pooling operation is critical so as to ensure permutation invariance. 
In the original PointNet~\cite{qi2017pointnet} paper, max and mean poolings are discussed. 
Bueno~\etal~\cite{bueno2021representation} explained the mechanism of pooling functions on sets and add sum pooling as an aggregation function. 
Pl-Net3D~\cite{mukhaimar2019pl} used deep declarative networks~\cite{gould2019deep} used M-estimators~\cite{rousseeuw2005robust} to provide additional robustness in pooling. However, due to its high computational cost and complexity it is not useful in most practical scenarios.
\cite{mukhaimar2021robust} and~\cite{soelch2019deep} explored more robust options for pooling functions like median, histogram (mode), and RANSAC (mode). In this paper, we propose to use PEs to replace the learned PPEs in PointNet and PCT to enable the use of robust pooling functions.

\section{Method}
\label{sec:method}

In this section, we provide a brief overview of point cloud processing architecture (\cref{fig:method}), which was proposed by PointNet/PCT. 
Suppose we have a 3D point cloud $\X\:{=}\:\{\x_1,\x_2,{\cdots},\x_N\}$ as input, where $\x_i \:{\in}\: \mathbb{R}^3$ and $N$ is the number of points.
The entire point cloud encoder computes a permutation invariant feature $\y \:{\in}\: \mathbb{R}^{K}$ induced by a mapping function $f\:{:}\: \mathbb{R}^{3 {\times} N} \:{\to}\: \mathbb{R}^{K}$, as described in~\cref{equ:point_cloud}.
\begin{equation}
    \label{equ:point_cloud}
    f\left(\X\right) = f\left(\{\x_1,\x_2,{\cdots},\x_N\}\right)\:.
\end{equation}

The mapping function $f$ can be divided into two parts: i) a per-point encoder $h\:{:}\:\mathbb{R}^{3} \:{\to}\: \mathbb{R}^K$ and ii) a symmetric (permutation invariant) aggregation function, usually max pooling. 
Thus, the complete point cloud encoder can be represented as:
\begin{equation}
\label{equ:perpointpool}
\y = \mbox{max}\left({h(\x_1),h(\x_2),{\cdots},h(\x_N)}\right)\:,
\end{equation}
For example, in PointNet~\cite{qi2017pointnet}, $h$ is learned mainly via MLPs; 
in naive PCT~\cite{guo2021pct}, $h$ is the attention layer; 
in addition, a complete PCT~\cite{guo2021pct} introduces MLP-based input pre-processing and sampling grouping layers to improve the performance of the naive PCT.
These three different types of $h$ directly operate on each point and can be viewed as per-point encoders.
After $h$, a max pooling operation is added to aggregate the final feature vector for the entire point cloud.
The entire network is trained end-to-end.=

However, a recent study~\cite{sanghi2020powerful} demonstrates that the deep ReLU PPE in PointNet does not necessarily need to be trained; instead, randomly initialized deep ReLU PPE performs well in point cloud classification tasks. 
This finding motivates us to explore other properties of the PPE beyond its learning capabilities. 
While randomness makes the encoder data-independent, which potentially helps its performance on data with OOD noise, properties of the PPEs and pooling options are also crucial to the OOD performance.

\subsection{Band-limited per-point encoder}
\label{sec:band-limited}

\begin{figure}[t]
    \centering
    {\includegraphics[width=0.9\linewidth]{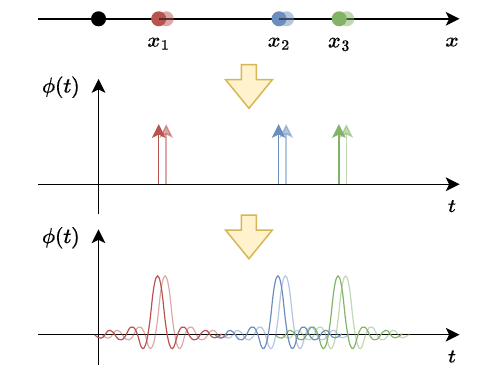} }%
    \vspace{-0.4cm}
    \caption{Illustration of a 1D point cloud and its embeddings.
    The upper row shows a simple 1D point cloud. 
    The middle row is the encoding with $\delta(\cdot)$. 
    The bottom row is the encoding with $\mbox{sinc}(\cdot)$.
    The faded ones indicate points with noise.}
    \label{fig:1D_encoding}%
    \vspace{-0.5cm}
\end{figure}

An obvious strategy to obtain a permutation invariant lossless feature is to encode points with the impulse function:
\begin{equation}
    \label{eq:impulse}
    \phi(t) = \sum_{i=1}^N h(x_i,t)= \sum_{i=1}^N \delta(t - x_i)\:,
\end{equation}
where $\delta(\cdot)$ is the impulse function and $t \:{\in}\: (0,1)$. 
A drawback is that $\phi(t)$ becomes an infinite bandwidth signal, which means we need infinite feature dimensions in discrete space to represent it (observe that this is equivalent to doing one-hot encoding in a continuous space). 
Another drawback is that when there is noise in the points, the embedding obtained by the impulse function will have a large difference from the embedding of the clean point cloud.
Additionally, the error between the clean point embedding and the noisy point embedding is not proportional to the noise intensity.
Nonetheless, a common approximation to the impulse function is $\mbox{sinc}(x) \:{=}\: \frac{\sin(\pi x)}{\pi x}$, which is band-limited and easy to discretize. 
The feature can be represented as:
\begin{equation}
    \label{eq:sinc}
    \phi(t) = \sum_{i=1}^N h(x_i,t)= \sum_{i=1}^N \mbox{sinc}(K(t-x_i))\:,
\end{equation}
where $K$ is a hyper-parameter that controls the bandwidth. 
\cref{fig:1D_encoding} provides a visual example. 
Note that shifting a signal does not affect its spectrum, and the Fourier transform is a linear operation. 
Therefore, the bandwidth of $\phi(t)$ is $K$, and it can be represented as $K$ equally spaced, weighted $\mbox{sinc}$ functions as,
\begin{equation}
    \label{eq:sinc_disc}
        \phi(t) =  \sum_{k=1}^K \gamma_k\mbox{sinc}\left(K\left(t-u_k\right)\right)\:,
\end{equation}
where $u_k$ are equally spaced (Nyquist Shannon sampling theorem). 
This can equivalently be represented in the Fourier domain as,
\begin{equation}
    \label{eq:sin_disc}
        \phi(t) = \sum_{b=0}^{B-1} \gamma_b\mbox{cos}\left(2\pi bt\right)+ \gamma_{b+B}\mbox{sin}\left(2\pi bt\right)\:,
\end{equation}
where $B \:{=}\: \frac{K}{2}$, and $\gamma$s are coefficients.

For high-dimensional cases, like 3D for point clouds, all the axes need to be sampled and encoded, and then the embeddings are multiplied together to get a high-dimensional embedding. The drawback of $\mbox{sinc}$ encoding is the huge number of bases required so that the encoding dimension is growing as a power to the input dimension. 
However, one huge advantage of sinusoids is that, although it seems no difference using $\mbox{sinc}$ or sinusoids basis in 1D, in high dimensional cases sinusoids can provide a better encoding with much fewer feature dimensions by trigonometric identities. 
Theoretically, the number of terms of $\mbox{sinc}$ or sinusoids basis in high dimension is the same. 
The distribution of these sinusoids is no longer uniform and tends towards a Gaussian whose variance is proportional to the bandwidth $B$. 
If one attempts to subsample from those huge numbers of basis, this non-uniform distribution offers a principled strategy for frequency selection. 
This gives some intuition on why random Fourier features (RFF)~\cite{tancik2020fourier} are such an attractive embedding for higher dimensional cases. 
We explain the high-dimensional case in the appendix and refer the reader to~\cite{zheng2022trading} where this dilemma is discussed at length.

\subsection{Smoothness of per-point embedding}
\label{sec:smoothness}

\begin{figure}[t]
    \centering
    \subfloat[\centering]{{\includegraphics[width=0.9\linewidth]{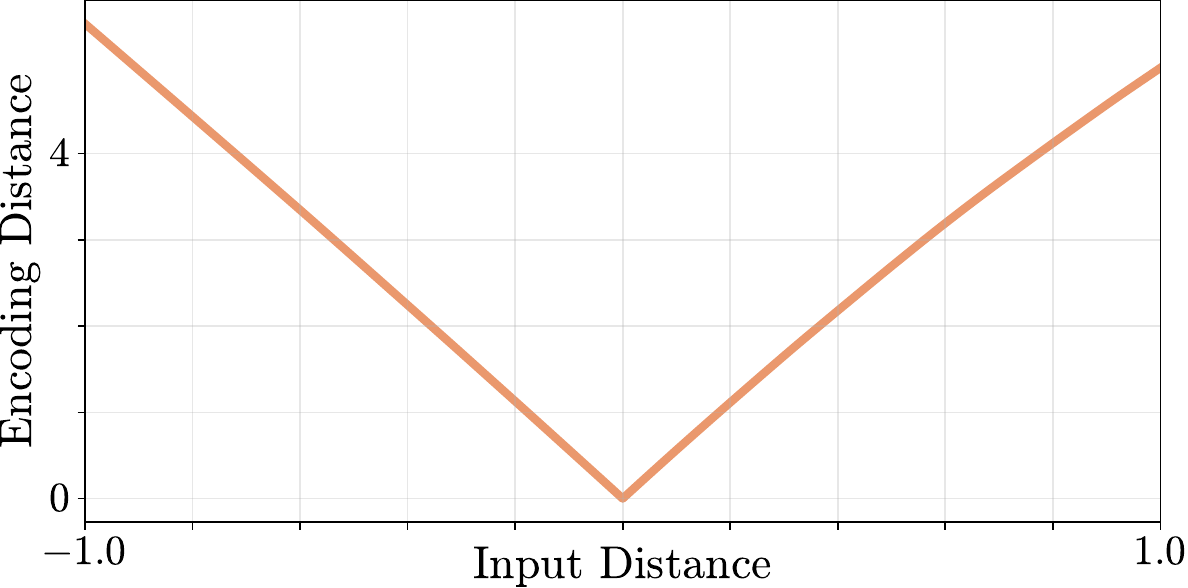} }}%
     \qquad
     \hspace{-1.5em}
    \subfloat[\centering]{{\includegraphics[width=0.9\linewidth]{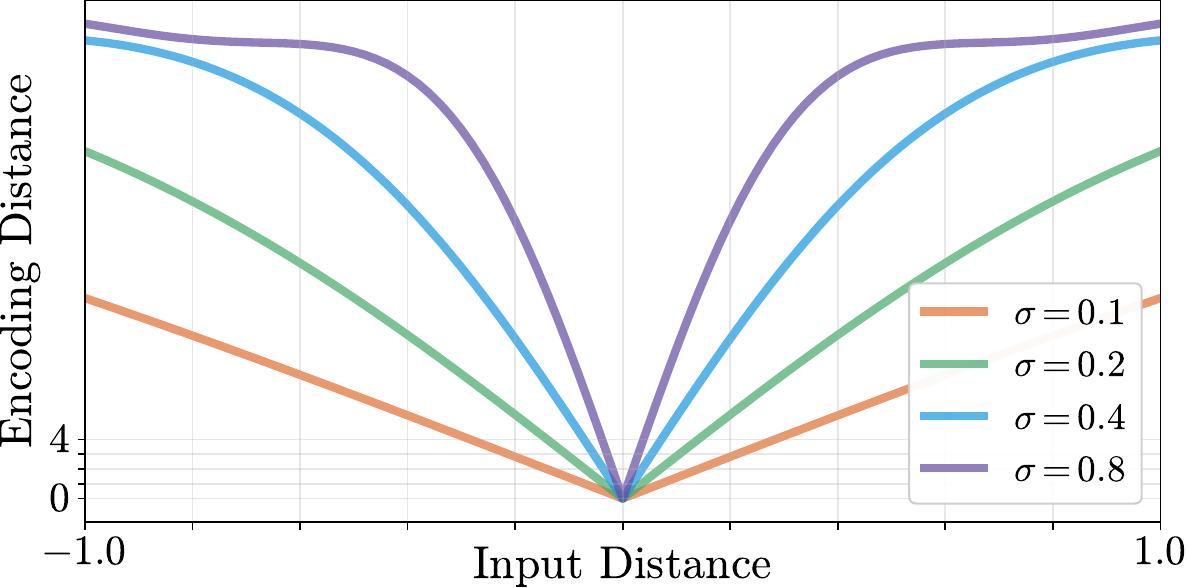} }}%
    \caption{Encoding distance \vs point distance for (a) one ReLU layer and (b) RFF-based PE. }%
    \label{fig:locality}%
    \vspace{-0.4cm}
\end{figure}

According to~\cite{zheng2021rethinking}, a good embedding should also be smooth (\ie, distance preserving), which means closer point clouds in the input space have more similar features. 
The feature is also important for the robustness of the model. 
When the input point cloud has noise or outliers, a smooth per-point encoder can generate a similar embedding to the clean one.
In the ideal case, feature distance should be proportional to the distance between input point clouds, which is
\begin{equation}
\label{eq:1D_smooth}
\begin{aligned}
\mathrm{D}_f(\y_{1},\y_{2})&\:{=}\:\|\phi(\{x_{1,i}\}_{i=1}^N,t){-}\phi(\{x_{2,i}\}_{i=1}^N,t)\|^2_2
\\&\:{\propto}\:\sum_{i=1}^{N}\|x_{1,i}{-}x_{2,i}\|_2^2\:,        
\end{aligned}
\end{equation}
where $\mathrm{D}_f$ is distance between features, $\{x_{1,i}\}_{i=1}^N$, $\{x_{2,i}\}_{i=1}^N$ are two point clouds and $\phi_1$, $\phi_2$ are their features.

When the PPE is not band-limited, such as the impulse function, this will not hold since no matter how far the point moves, the changes in the feature are the same. 
When the PPE is band-limited as we discussed in~\cref{sec:band-limited}, based on the theory in~\cite{zheng2021rethinking}, for each point, the encoding distance is proportional to the input distance, which means
\begin{equation}
    \label{eq:distance_preservation}
        \mathrm{D}_e(\e_{1},\e_{2})=\|\phi(x_1,t){-}\phi_2(x_2,t)\|^2_2\propto\|x_{1}{-}x_{2}\|_2^2\:,
\end{equation}
where $\mathrm{D}_e$ means distance between encodings, $x_1$, $x_2$ are two point clouds and $\e_1$, $\e_2$ are their features. 
Although it is not necessarily equivalent, it strongly implies~\cref{eq:1D_smooth} will hold since features are just summation of encodings. 
In addition, a critical advantage of using our band-limited encoders is that one can easily tune the bandwidth of the representation by varying the variance of the frequency distribution. 
\cite{zheng2021rethinking} shows that by varying the bandwidth, the distance preservation (or the feature similarity) between nearby point encodings can be controlled. 
This is a vital property that can be exploited when dealing with OOD data.

\subsection{Trade-off -- bandwidth and smoothness}
\label{sec:robust}
In the original PointNet architecture, the encoder is an MLP and the entire network is trained end-to-end. 
The deep ReLU point embedding for PointNet can be defined as
\begin{equation}
  \mathcal{H}_{\ell}(\mathbf{x}) = \mbox{ReLU}[\mathbf{W}_{\ell} \circ
  \mathcal{H}_{\ell - 1}(\mathbf{x}) + \mathbf{b}_{\ell-1}]\:,
\end{equation}
where $\mathcal{H}_{0}(\mathbf{x}) \:{=}\: \mathbf{x}$. 
The PPE in PCT can be defined as
\begin{equation}
\begin{aligned}
    \mathbf{F}_1 &= \mbox{AT}^{1}(\mathbf{F}_{e})\:, \notag\\
    \mathbf{F}_i &= \mbox{AT}^{i}(\mathbf{F}_{i-1})\:,\:i=2,3,4 \notag \\
    \mathbf{F}_o &= \mbox{concat} (\mathbf{F}_{1},\mathbf{F}_{2},\mathbf{F}_{3},\mathbf{F}_{4})\:,
\end{aligned}
\end{equation}
where $\mbox{AT}^{i}$ is the $i$-th attention layer, $\mathbf{F}_{e}$ is the input embedding through some MLPs and sampling and grouping layer, and $\mathbf{F}_o$ is the final embedding. 
The deep ReLU network and attention layers map low-dimensional encodings to a high-dimensional space. 
This mapping is band-limited and preserves distance information when randomly initialized. 
However, after training end-to-end with data, its ability to preserve distance entirely depends on the training data. 
There is no clear geometry information in the learned features, therefore, when OOD noise is added to the data, it is unpredictable how the features will change in the feature space and whether it will pass the decision boundary of the learned classifier or not.

Moreover, even though random initialized MLPs can be viewed as a non-trained per-point encoder, it is unclear how the initialization affects the behavior of the embeddings. 
On the contrary, PE has a well-established theory---by tuning the hyper-parameters (usually called ``scale'' in PE), we can theoretically know the bandwidth and smoothness changes and predict the performance of the embedding. 
The sinusoids encoder we discussed in~\cref{sec:band-limited} turns into PE---RFF, which is defined as
\begin{equation}
  \mathcal{G}(\mathbf{x}) = \begin{bmatrix} \cos(\mathbf{W}
    \mathbf{x}) \\ \sin(\mathbf{W}
    \mathbf{x})  \end{bmatrix}\:.
 \end{equation}

The distance preservation property of these three encoders is shown in~\cref{fig:locality}. 
For the ReLU network, the distance is kept linearly in the point space, while for RFF, this property can be tuned by tuning the scale (standard deviation of random weight matrix). 
Even if we change the standard deviation of the ReLU network,~\ie, multiply the weights with a specific scale, we will get an output that is also multiplied by that specific scale, instead of changing the shape of the distance preservation curve. 
Note that for RFF, when the scale is large, such as $\sigma{=}8$ (shown in~\cref{fig:locality}), only local distance is preserved, which means only local information is concerned. 
This property makes the RFF encoder robust to noise and outliers.

\section{Experiments}
\label{sec:exp}

\begin{table}[t]
\caption[]{Classification accuracy(\%) on ModelNet40 of RFF-based PE with different scales of RFF.}
\centering
\begin{adjustbox}{width=\linewidth}
\begin{tabular}{lcccccccc}
\toprule
\thead{\normalsize Scale} & \thead{\normalsize 0.05} & \thead{\normalsize 0.1} & \thead{\normalsize 0.5} & \thead{\normalsize 1} & \thead{\normalsize 5} & \thead{\normalsize 10} & \thead{\normalsize 15} & \thead{\normalsize 20} \\
\midrule
Train & 80.9 & 82.8 & 95.6 & 99.2 & 100 & 100 & 100 & 100 \\
Test & 66.2 & 79.2 & 86.1 & 86.1 & 76.1 & 33.8 & 23.0 & 19.6 \\
\bottomrule
\end{tabular}
\label{tab:scale_cls}
\end{adjustbox}
\vspace{-0.4cm}
\end{table}

\begin{table*}[t]
\caption[]{Classification results of different models on modified ModelNet40-C (generated by larger corruption severities). 
All results shown here are classification errors. 
The upper tabular between {\color{sol_blue}\textbf{blue bars}} are methods with end-to-end training. 
The lower tabular between {\color{beer_orange}\textbf{orange bars}} are methods with randomly initialized PPEs and trained classifiers.
All the training is on ModelNet40 clean data. }
\centering
\begin{adjustbox}{width=\textwidth}
\begin{tabular}{lccccccccccccc}
\toprule
\multirow{2}{*}{\thead{\normalsize Method}} 
& \multirow{2}{*}{\thead{\normalsize Clean}} 
& \multirow{2}{*}{\thead{\normalsize Corruption}} 
& \multicolumn{3}{c}{\thead{\normalsize Noise Corruption}}
&& \multicolumn{5}{c}{\thead{\normalsize Outliers Corruption}} \\
\cmidrule{4-6} \cmidrule{8-12}
&    & (avg.)   & Uniform       & Gaussian      & Impulse       
&& Upsampling    & Background    & Ball(10\%)    & Ball(50\%)    & Ball(100\%)   \\ \midrule
\arrayrulecolor{sol_blue}\toprule[0.3ex]
PointNet~\cite{qi2017pointnet} & 11.3 & 55.8 & 18.3 & 41.4 & 28.5 
&& 13.4 & 94.7 & 82.5 & 83.7 & 83.9 \\
PointNet++~\cite{qi2017pointnet++} & \textbf{7.5} & 43.8 & 37.3 & 70.1 & 47.2 
&& \underline{12.4} & 49.1 & 27.1 & 48.3 & 58.8 \\
DGCNN~\cite{wang2019dynamic} & \underline{8.1} & 52.7  & 35.2 & 67.0 & 35.7          
&& 21.9 & 66.5 & 50.9 & 44.6 & --- \\
RSCNN~\cite{liu2019relation}  & 9.3 & 52.7 & 30.4 & 65.5 & 49.9 
&& \textbf{10.5} & 60.6 & 54.6 & 75.3 & 74.4 \\
PCT~\cite{guo2021pct} & \underline{8.1} & 51.9 & 22.2 & 57.8 & 37.1
&& \underline{12.4} & 71.5 & 61.2 & 75.0 & 77.9 \\
\arrayrulecolor{sol_blue}\toprule[0.3ex]
\arrayrulecolor{beer_orange}\toprule[0.3ex]
PE-AT (Ours) & 15.6 & \underline{23.3} & \underline{15.1} & \textbf{17.1} & \underline{16.0}
&& 14.7 & \underline{19.4} & \underline{16.0} & \underline{33.4} & \underline{55.1} \\
PE (Ours) & 13.6 & \textbf{22.1} & \textbf{14.1} & \textbf{17.1} & \textbf{15.7}
&& 13.7 & \textbf{17.9} & \textbf{14.6} & \textbf{31.8} & \textbf{52.3} \\
\arrayrulecolor{beer_orange}\toprule[0.3ex]
\arrayrulecolor{black}\bottomrule
\end{tabular}
\end{adjustbox}
\label{tab:robust}
\end{table*}

In this section, we first show that PE can replace PPE with classification results on the clean ModelNet40~\cite{wu20153d} dataset and have the ability to control bandwidth by tuning the scale of RFF-based PE. 
Then we use our modified ModelNet40-C~\cite{sun2022benchmarking} dataset that contains various corruptions to show that our PE and PE-AT methods are much more robust than current end-to-end trained methods when dealing with noisy data.
We show the performance of using different pooling functions to validate our theory of how to get a robust model.
Finally, we showcase point cloud registration results on ModelNet40 in OOD scenarios.

\subsection{Experimental settings}
As mentioned before, the architecture of PointNet and PCT can be divided into three parts,~\ie, a PPE (ReLU-MLPs), a pooling function (max pooling), and a downstream network (fully connected networks, FCNs).
Here, we introduce our settings for these three components separately.

\vspace{0.15cm}
\noindent\textbf{Per-point embeddings.}
We used two different types of PPEs.
We denote the attention layer as AT, the linear layer as FC (fully connected), the ReLU activation function as ReLU, and the random Fourier features as RFF.
The dimension of the input and output feature to the encoder is denoted as (X, Y).
\textbf{PE-AT (RFF-based PE + attention)} is a variation of PCT. 
We use PE as the input embedding and directly feed the embeddings into 4 attention layers: RFF (3, 256) $\rightarrow$ AT.
The scale of RFF is $0.9$.
\textbf{PE (RFF-based PE)} uses PE as the surrogate PPE: RFF (3, 1024).
The scale of RFF for max pooling is $0.09$, and for all the other pooling functions (mean/median), we used $0.9$.

\vspace{0.15cm}
\noindent\textbf{Pooling functions.}
We used three types of pooling operations: max pooling, mean pooling, and median pooling.
Max pooling is widely used in current models, which results in poor performance in OOD cases.
Mean pooing is used in our models to maintain robustness to noise.
Median pooling is used in the noisy data experiments to further demonstrate that PE enables a more robust choice of pooling functions that better deal with OOD noise.

\vspace{0.15cm}
\noindent\textbf{Downstream networks.}
For the classifier task, we used the exact network in the original PointNet~\cite{qi2017pointnet}:
FC (1024, 512) $\rightarrow$ BN $\rightarrow$ ReLU $\rightarrow$ FC (512, 256)$ \rightarrow$ Dropout (0.4) $\rightarrow$ BN $\rightarrow$ ReLU $\rightarrow$ FC (256, \#class).
Here, Dropout (0.4) is the dropout layer with 40\% of the probability of an element being zero and BN refers to the batch normalization layer.
The number of object categories in classification (\#class) is 40. 
Note that the classifier in the original PCT is different. 
Here we changed the classifier in PCT to be the same as the one used in PointNet for a fair comparison. 
Note that the modification to the classifier does not affect the performance of the method.

\subsection{Classification on clean data}
We followed the standard training protocol in~\cite{sun2022benchmarking} and trained all models on clean ModelNet40~\cite{wu20153d} data with $1024$ points and $300$ epochs. 
We did not use any data augmentation strategy except random scaling and translation as proposed in DGCNN~\cite{wang2019dynamic}---which is usually considered as ``unaugmented''. 
We show the classification results on clean data in the first columns of~\cref{tab:robust}.
We observed that end-to-end trained methods achieve good performance and PointNet++ has the lowest error rate.
Both randomly initialized PE-AT and PE achieve comparable but slightly worse results compared to learned PointNet indicating the ability of PE-based methods in replacing learned PPEs.

\vspace{-0.2cm}
\paragraph{Ablation study on different scales of PE}
As mentioned in~\cref{sec:smoothness}, different scales (standard deviations) decide the bandwidth and distance preservation of RFF-based PE, therefore affecting the performance. 
We demonstrate this behavior in~\cref{tab:scale_cls}, where increasing the standard deviation in RFF leads to higher training accuracy but lower testing accuracy beyond a certain threshold. 
This property also applies to all other PEs, such as the Gaussian encoder proposed in~\cite{zheng2021rethinking}.
We show additional results with Gaussian-based PE in the supplementary material.

\subsection{Robustness to noisy data}
\label{sc:robust_classify}
To compare the robustness of different PPEs and pooling functions, we trained the model (the entire model or the classifier alone if PPEs do not need to train) on clean data and tested these trained models on test data under six different types of noise and outlier corruptions.
We selected five noise/outlier types from ModelNet40-C~\cite{sun2022benchmarking} dataset: uniform noise, Gaussian noise, impulse noise, upsampling outliers, and background outliers.
However, the corruption generated by ModelNet40-C is too small, which makes the dataset less challenging. 
Therefore, we enhanced the severity of the corruption by either enlarging the variance of the noise or increasing the number of outliers adding to the clean point cloud.
We also included a ball outlier to better demonstrate the effectiveness and robustness of PE-based methods.
The detailed settings are listed below.

\textbf{Noise corruption} is a type of data corruption where the total number of points in a point cloud is unchanged, but a subset of points is corrupted by noise.
For example, a part of or all points are shifted from their original positions by noise.
Usually, this noise can be represented by a random distribution, such as uniform, Gaussian, and impulse. 
The severity range of the noise we used in the experiment is as follows.
\noindent \textbf{1) Uniform noise:} 10 different severity levels ranging from $0.01$ to $0.1$ were used for all points. 
\noindent \textbf{2) Gaussian noise:} a mean $0$ and standard deviation ranging from $0.01$ to $0.1$ Gaussian noise was added to all points. 
\noindent \textbf{3) Impulse noise:} an impulse with $0.1$ magnitude, $[1,1,1]$ or $[-1,-1,-1]$ direction was added to $2\%$ to $100\%$ of points.

\textbf{Outlier corruption} is a type of data corruption where a small portion of additional points was added to the clean point cloud, yielding an increase in the total number of points.
We used three different types of outliers,~\ie, upsampling, background, and ball outliers.
We list the severity parameters of the outliers we added as follows.
\textbf{4) Upsampling outliers:} $10\%$ to $100\%$ of points were randomly copied and modified with uniform noise in $[-0.05,0.05]$. 
Then these points were added to the original point cloud as outliers.
\textbf{5) Background outliers:} $2\%$ to $100\%$ of points sampled uniformly from $[-1,1]$ were added as additional points.
\textbf{6) Ball outliers:} we designed a ball outlier that was uniformly sampled on a ball surface with a radius ranging from $0.3$ to $3.0$. 
We added $10\%$, $50\%$, and $100\%$ of points as additional points.

We show the performance of our methods compared to other learned PPEs under different data corruptions in~\cref{tab:robust}.
Our methods achieve compelling results across all data corruption types, indicating the great advantage of using untrained PE as a surrogate for learned PPEs under OOD scenarios.
Specifically, our methods maintain great accuracy yielding a substantial decrease in error rate when common corruptions like Gaussian noise and background outliers are presented in the data.
The superior performance of our methods further validates the robustness of PE-based embeddings in dealing with OOD cases such as noise and outliers.
Moreover, the bandwidth of PE could be easily tuned through the scale of the embedding to control the ability to handle high-level noise corruption.

\begin{figure}[t]
    \centering
    {\includegraphics[width=0.9\linewidth]{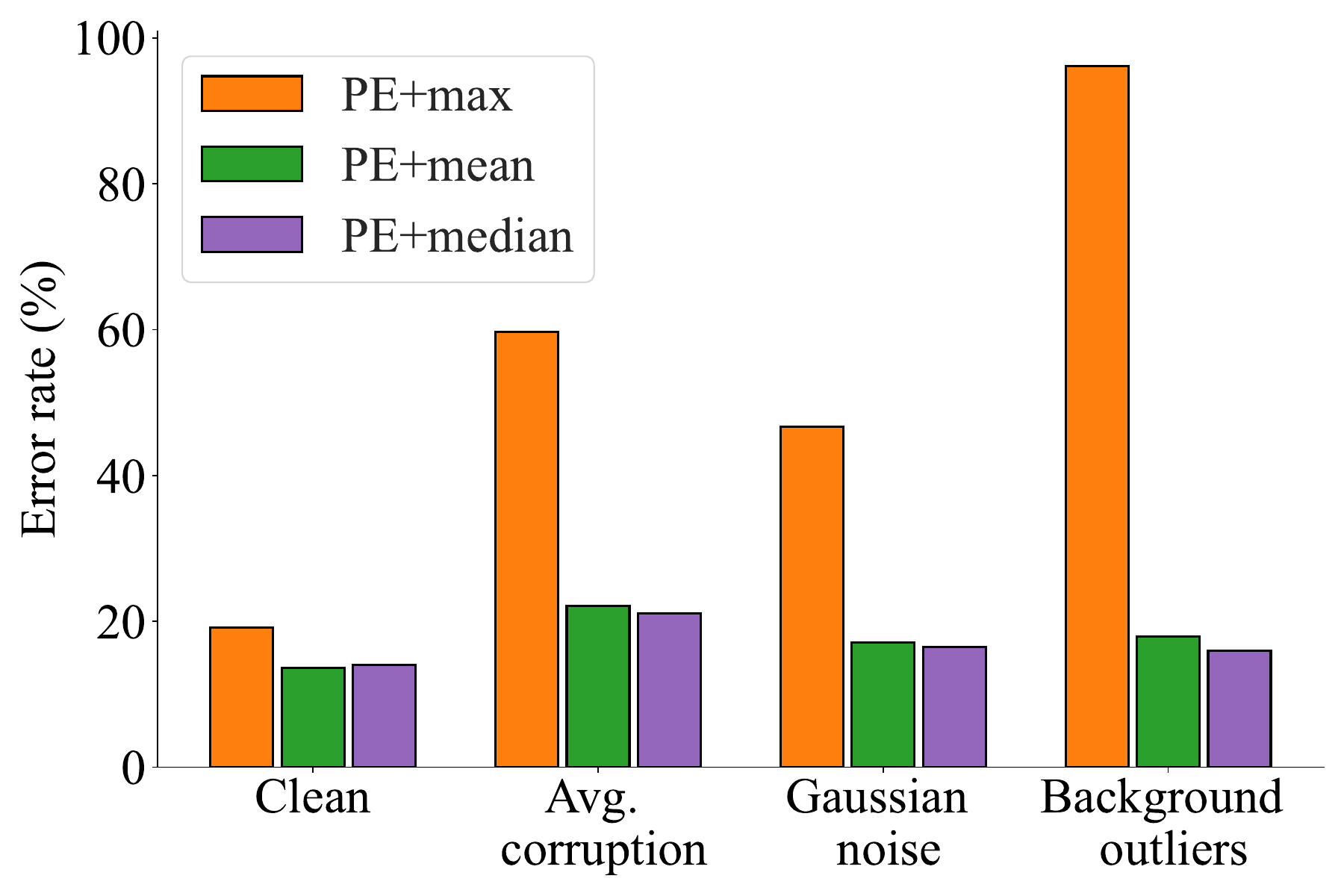} }%
    \caption{Error rate of our PE method with different pooling functions under different data corruptions.}%
    \label{fig:pooling_ablation}%
    \vspace{-0.3cm}
\end{figure}

Our methods achieve state-of-the-art results on almost all corruption types.
However, in the upsampling outlier corruption case, all methods (except DGCNN) achieve similarly good performance.
The reason behind this is that the point outliers were generated from upsampled points with a very small uniform noise ranging in $[-0.05,0.05]$ that does not account for a strong OOD case.
For other outlier corruption experiments with uniformly sampled background outliers and outliers sampled on a ball surface, PointNet~\cite{qi2017pointnet} fails catastrophically due to its PPE processing each point independently and only contains global features, making the point embedding easily affected by these outliers.
Other learned PPEs~\cite{qi2017pointnet++, wang2019dynamic, liu2019relation, guo2021pct} combine local information in neighboring points that help the resistance to small outliers.
Nevertheless, these local patterns are still biased to the training data and cannot achieve robust performance when handling large noise/outlier corruptions.
We have also noticed that the randomly initialized attention (PE-AT) will not harm the performance of the untrained PE (PE-AT) and achieve comparable results to PE.
We speculate that the relation map between points in a point cloud is not biased on the training data.

\begin{table}[t]
\caption[]{Ablation study on different architectures of PCT and PE-AT. 
The upper tabular methods between {\color{sol_blue}\textbf{blue bars}} use trained end-to-end PPEs, and the lower tabular methods between {\color{beer_orange}\textbf{orange bars}} use randomly initialized PPEs.}
\centering
\begin{adjustbox}{width=0.8\linewidth}
\begin{tabular}{lccc}
\toprule
Model       & Pooling & Clean ER   & Corruption ER             \\ 
\midrule
\arrayrulecolor{sol_blue}\toprule[0.3ex]
PCT~\cite{guo2021pct}   & max       & \textbf{8.1}                       & 51.9               \\
PCT~\cite{guo2021pct}  & mean      & 8.8                       & 52.7            \\
PE-AT (Ours) & max       & 10.1                       & 47.7               \\
PE-AT (Ours) & mean      & 11.2                       & 37.7                \\ 
\arrayrulecolor{sol_blue}\toprule[0.3ex]
\arrayrulecolor{beer_orange}\toprule[0.3ex]
PCT~\cite{guo2021pct} & max       & 14.3                       & 55.2               \\
PCT~\cite{guo2021pct} & mean     & 13.0                       & 49.4                 \\
PE-AT (Ours)   & max        & 17.3                       & 39.3                \\
PE-AT (Ours)   & mean     & 15.6                       & \textbf{23.3} \\ 
\arrayrulecolor{beer_orange}\toprule[0.3ex]
\arrayrulecolor{black}\bottomrule
\end{tabular}
\end{adjustbox}
\label{tab:pct}
\vspace{-0.2cm}
\end{table}

\subsection{Ablation study of different pooling functions}
We show the performance of our PE embedding on different corrupted data using different pooling functions, such as max, mean, and median in~\cref{fig:pooling_ablation}.
On clean data, all pooling functions achieve similarly good performance, and mean/median poolings have slightly better accuracy.
However, when the data is corrupted, for example, Gaussian noise or background outliers are added, max pooling yields catastrophic errors while mean and median poolings maintain robust and good accuracy.

\subsection{Ablation study of PCT and PE-AT}
In~\cref{tab:pct}, we show the results under different PPEs (end-to-end trained or randomly initialized) and pooling (max or mean) settings of PCT and PE-AT using both clean and corrupted data.
On the clean data, all methods using mean pooling show slightly worse performance than using max pooling when doing end-to-end training, since max pooling allows for more powerful feature learning.
On the contrary, when using randomly initialized PPEs, all methods using mean pooling result in slightly better accuracy due to the data-unbiased nature of these random PPEs.
For corruption cases, if we only change one condition from the original model (PCT+Max, $51.9\%$), changing max pooling to mean pooling (PCT+Mean, $52.7\%$) or just using randomly initialized per-point encoder (PCT(R)+Max, $55.2\%$) will not work. 
However, only changing the input embedding to PE (PE-AT+Max, $47.7\%$) works slightly better. 
When two conditions are changed, the one without PE (PCT(R)+Mean, $49.4\%$) is the worst. 
The ones with PE (PE-AT+Mean, $37.7\%$ \& PE-AT(R)+Max, $39.3\%$) already show good robustness on OOD noise and outliers. 
Finally, if we change all three conditions (PE-AT(R)+Mean, $23.3\%$), the model works best on corruption cases.

\begin{figure}[t]
    \centering
    {\includegraphics[width=0.9\linewidth]{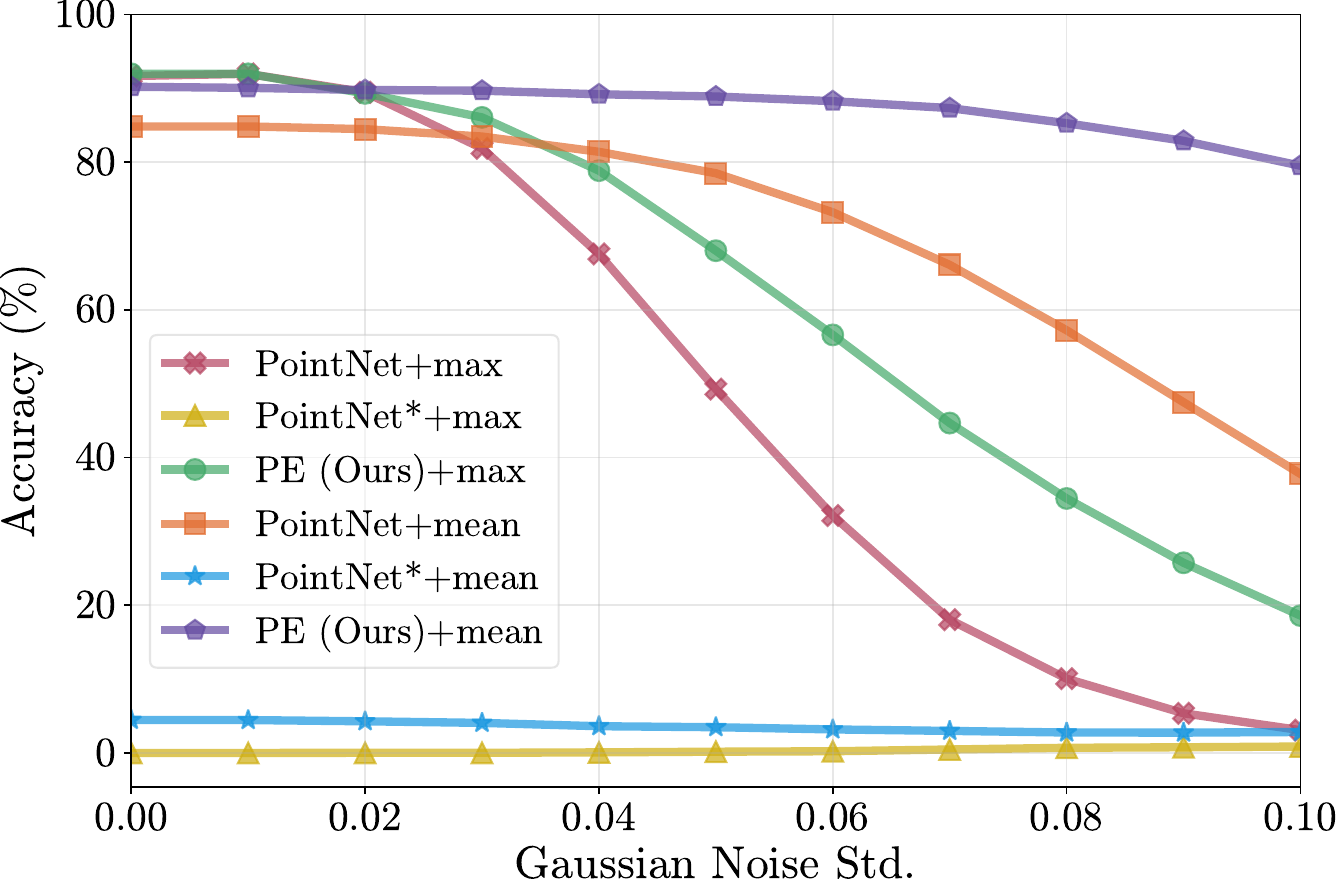} }%
    \caption{Point cloud registration accuracy on ModelNet40 20 categories with Gaussian noise.
    Our method achieves robust registration results on noisy data, especially with mean pooling.
    All methods use the Lucas-Kanade algorithm for alignment.
    Note that PointNet* refers to PointNet with randomly initialized PPE.}%
    \label{fig:alignment_noise}%
\end{figure}

\subsection{Point cloud registration}
We test the robustness of different pooling functions and different PPEs (PointNet or PE) on the point cloud registration task using the ModelNet40 dataset.

We chose the point cloud registration method~\cite{li2021pointnetlk} that uses a simple PointNet structure---3-layer MLPs---as a feature encoder as our baseline.
In PointNetLK Revisited~\cite{li2021pointnetlk}, a point-based Lucas-Kanade algorithm is used for registration, which is essentially a Gauss-Newton method.
Note that the learning-based method was only trained on a clean ModelNet dataset.
We then replaced the PPE in PointNet with PE and kept the Gauss-Newton registration algorithm as our method.

\begin{figure}[t]
    \centering
    \includegraphics[width=\linewidth]{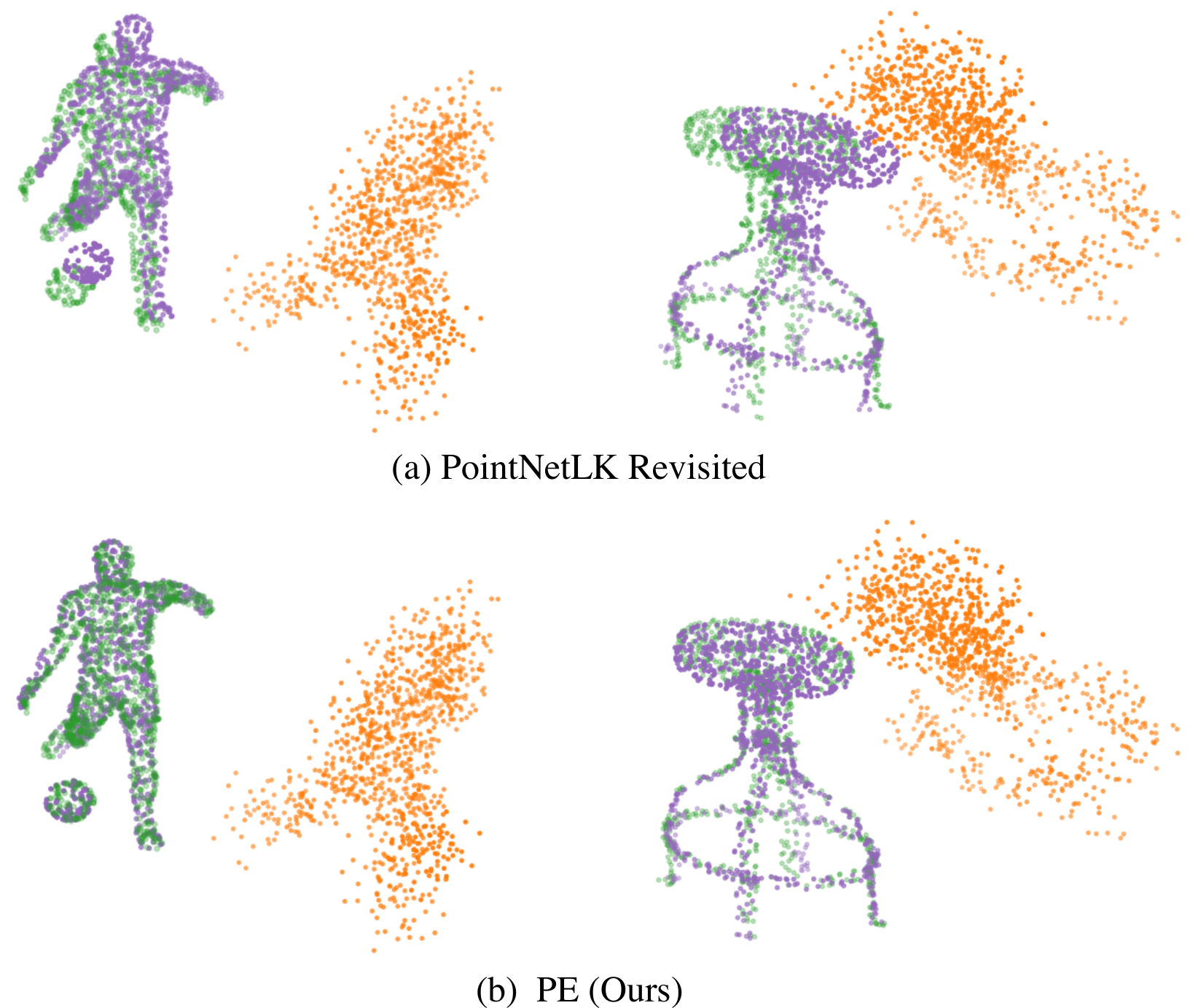}
    \caption{A visualization of point cloud registration using (a) PointNetLK Revisited~\cite{li2021pointnetlk} and (b) PE on ModelNet40 dataset. 
    Gaussian noise with a standard deviation of 0.05 is presented.
    Green: clean target points, orange: transformed source points with Gaussian noise, and purple: source points transformed by estimated pose. 
    Note that we only show purple points in clean data to better visualize the shape of the object.
    In the experiments, we used noisy data as shown in orange.
    The learning-based method with the PointNet encoder cannot properly handle noise, leaving a large discrepancy between the deformed source point cloud and the target point cloud.
    In contrast, our PE method successfully aligns two point clouds even when large noise is added.}
    \label{fig:clear_reg}%
\end{figure}

Similar to~\cref{sc:robust_classify}, we added Gaussian noise with a zero mean and a standard deviation ranging from 0.01 to 0.1 with step size 0.01. 
As shown in~\cref{fig:alignment_noise}, PointNetLK Revisited~\cite{li2021pointnetlk} achieves good performance when there is only small noise presented.
However, when large noise is added, the performance drops dramatically, especially for untrained PPE.
In contrast, our method (PE) maintains good performance even when large noise is presented.
Note that mean pooling yields the most robust performance indicating the effectiveness of our method.
The visual comparison is shown in~\cref{fig:clear_reg}.

\begin{table}[t]
\caption[]{Performance on other data corruptions as provided in ModelNet-C~\cite{sun2022benchmarking}.
We showcased our performance using increased density, cutout density, rotations, and shear transformations.}
\centering
\begin{adjustbox}{width=0.8\linewidth}
\begin{tabular}{lcccc}
\toprule
Model       & Density Inc. & Cutout   & Rotation & Shear            \\ 
\midrule
\arrayrulecolor{sol_blue}\toprule[0.3ex]
PointNet   & \textbf{11.7} & \underline{13.2}                       & 34.8 &25.8              \\
PointNet++  & 17.5      & \textbf{11.6}                 &27.6      & \underline{15.0}           \\
PCT & \underline{11.8}       & 13.3                       & \textbf{16.4} & \textbf{10.8}               \\
\arrayrulecolor{sol_blue}\toprule[0.3ex]
\arrayrulecolor{beer_orange}\toprule[0.3ex]
PE-AT (Ours) & 20.1     & 19.0                       & 36.3 &31.5                 \\
PE (Ours)   & 18.3        & 17.3                       & \underline{27.1} &22.3                \\
\arrayrulecolor{beer_orange}\toprule[0.3ex]
\arrayrulecolor{black}\bottomrule
\end{tabular}
\end{adjustbox}
\label{tab:limitation}
\end{table}

\vspace{0.2cm}
\section{Limitations}

Our proposed PE-based PPEs have shown compelling robustness on noise and outlier corruptions. 
However, our method does not show superior advantages over learned PPEs with corruptions like density changes and transformations (see~\cref{tab:limitation}).
We acknowledge that this is due to the limitations of the RFF-based PEs.
For point density changes, a more specialized PE function is expected.
For transformation corruptions, pre-processed data normalization is needed.
The full experiments and analysis are provided in the supplementary material.
Nevertheless, the focus of the paper is to demonstrate that untrained PE is a robust point cloud embedding for OOD noise presented in data.

\section{Conclusion}

In this paper, we theoretically and empirically show that an untrained positional embedding can be a robust surrogate for learned per-point embedding in point cloud processing.
Moreover, we show that the per-point embeddings should be both band-limited and smooth to maintain good robustness to OOD corruptions where noise and outliers are presented in the data.
One of the advantages of using these untrained PPEs is that the bandwidth and the smoothness of the embedding can be manually tuned to control the resistance to the OOD noise.
We further demonstrate the downstream tasks such as object classification and point cloud registration through thorough experiments.


\bibliography{main}

\clearpage
\setcounter{page}{1}
\maketitlesupplementary

We provide additional details of our method and experiments in the supplementary material.

\section{RFF Derivation}
\label{app:theo}
\label{app:main_purpose}
In this section, we extend the derivation of Random Fourier Features (RFF) from band-limited embeddings discussed in the main text Sec.~{\textcolor{red}{3.1}}. 
In particular,
\begin{enumerate}
    \item We derive a per-point embedding (PPE) not trained on data that is better for out-of-domain (OOD) generalization.
    \item Empirically, sinusoidal embeddings are better than ReLU embeddings under randomly initialized conditions.
    This finding can be explained by the tradeoff between the controllable bandwidth and the local smoothness.
    \item We show why RFF is a band-limited PE and derive RFF from sinusoidal embeddings.
    Since RFF is a derivative of sinusoidal embeddings, it explains why it is a more robust PPE than other band-limited PEs.
\end{enumerate}

\subsection{RFF in 2D}
\label{app:2d_der}

\begin{figure}[t]
    \centering
    {\includegraphics[width=0.9\linewidth]{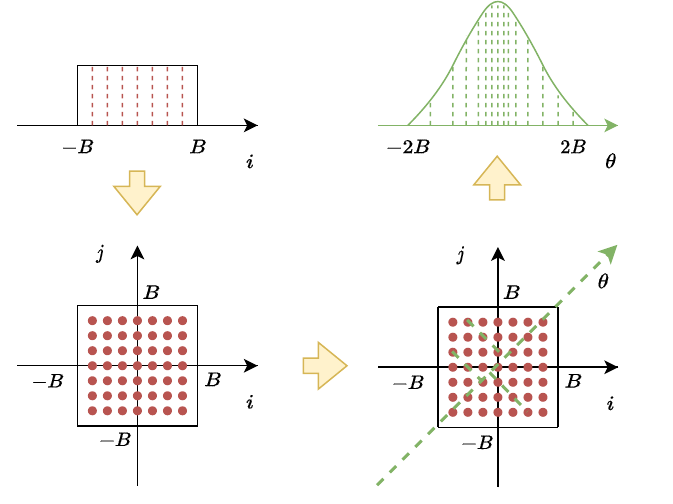} }%
    \caption{Illustration of how 1D and 2D band-limited embeddings sample the frequencies.
    The upper left corner depicts a histogram of coefficients for sinusoidal embeddings in 1D where frequencies are equally spaced between~$-B$ to~$B$ denoting the bandwidth. 
    For the 2D case (see~\cref{eq:sin_disc_2d}), a histogram of these coefficients can be seen in the bottom left corner where the frequencies are equally spaced in 2D. 
    The trigonometric identities used in~\cref{eq:trig_identity} are equivalent to the rotation of a 2D grid onto a new axis (see bottom right corner). 
    In the top right corner, one can see how the position of these frequencies on this rotated axis now tends towards a Gaussian---no longer a uniform distribution. 
    Note that there are still $K^2$ green dash lines on the $\theta$ axis, which means there are still $K^2$ different frequencies in 2D. 
    However, the frequency density shape gives us an idea of how to sub-sample the frequencies.
    }%
    \label{fig:sample_freq}%
\end{figure}

We show that for the 2D case, instead of using a $\mbox{sinc}$ function, we can use sinusoidal embeddings. 
Both of these embeddings form orthonormal bases in a $K$-dimensional space.

We want to emphasize that different encodings are equivalent to coordinates in different coordinate systems. 
For example, in the 2D case, the $K$ band-limited embedding by $\mbox{sinc}$ can be viewed as
\begin{equation}
    \label{eq:sinc_disc_2d}
        \phi(u,v) \:{=}\: \sum_{i=1}^K \sum_{j=1}^K\gamma_{i,j}\mbox{sinc}\left(K\left(u \:{-}\: x_i\right)\right)\mbox{sinc}\left(K\left(v \:{-}\: y_i\right)\right)\:,
\end{equation}
which requires $K^{2}$ samples for each combination of $x_i$ and $y_i$. 
Therefore, the sinusoidal embeddings can be represented as
\begin{align}
    \label{eq:sin_disc_2d}
    \phi(u,v) &= \sum_{i=0}^{B-1}\sum_{j=0}^{B-1} A_{i,j}\mbox{cos}\left(2\pi iu\right)\mbox{cos}\left(2\pi jv\right) \notag\\
            &+\sum_{i=0}^{B-1}\sum_{j=0}^{B-1} A_{i+B,j}\mbox{sin}\left(2\pi iu\right)\mbox{cos}\left(2\pi jv\right) \notag\\
            &+\sum_{i=0}^{B-1}\sum_{j=0}^{B-1} A_{i,j+B}\mbox{cos}\left(2\pi iu\right)\mbox{sin}\left(2\pi jv\right) \notag\\
            &+\sum_{i=0}^{B-1}\sum_{j=0}^{B-1} A_{i+B,j+B}\mbox{sin}\left(2\pi iu\right)\mbox{sin}\left(2\pi jv\right)\:,
\end{align}
which needs $K^{2}$ terms for each combination of $i$ and $j$. 
Following trigonometric identities
\begin{align}
    \label{eq:trig_identity}
    \sin{\alpha}\cos{\beta}&=\frac{\sin{(\alpha+\beta)}+\sin{(\alpha-\beta)}}{2} \notag\\
    \cos{\alpha}\cos{\beta}&=\frac{\cos{(\alpha+\beta)}+\cos{(\alpha-\beta)}}{2} \notag\\
    \sin{\alpha}\sin{\beta}&=\frac{\cos{(\alpha-\beta)}-\cos{(\alpha+\beta)}}{2},
\end{align}
we have,
\begin{equation}
    \label{eq:sin_disc_2d_t}
    \begin{aligned}
    \phi(u,v) &= \sum_{i=0}^{B-1}\sum_{j=0}^{B-1} C_{i,j}\mbox{cos}\left(2\pi(iu+jv)\right)
            \\&+\sum_{i=0}^{B-1}\sum_{j=0}^{B-1} C_{i+B,j}\mbox{cos}\left(2\pi(iu-jv)\right)
            \\&+\sum_{i=0}^{B-1}\sum_{j=0}^{B-1} C_{i,j+B}\mbox{sin}\left(2\pi(iu+jv)\right)
            \\&+\sum_{i=0}^{B-1}\sum_{j=0}^{B-1} C_{i+B,j+B}\mbox{sin}\left(2\pi(iu-jv)\right)\:.
    \end{aligned}
\end{equation}
Note that~\cref{eq:sin_disc_2d_t} and~\cref{eq:sin_disc_2d} are mathematically equivalent to each other. 
The coefficients $C$ and $A$ are coordinates under different coordinate systems. 
Although there are still $K^2$ terms, the transformation is not orthonormal.
The distributions of these sinusoids are no longer uniform but in the triangle shape of width $4B$---which is the sum of two uniform distributions on $[-B,B]$. 
If one attempts to sub-sample from these $K^2$ frequencies, this non-uniform distribution offers a principled strategy for frequency selection. 
Please see~\cref{fig:sample_freq} (a) for an example.

\begin{figure}[t]
    \centering
    \subfloat[\centering 2D]{{\includegraphics[width=0.4\linewidth]{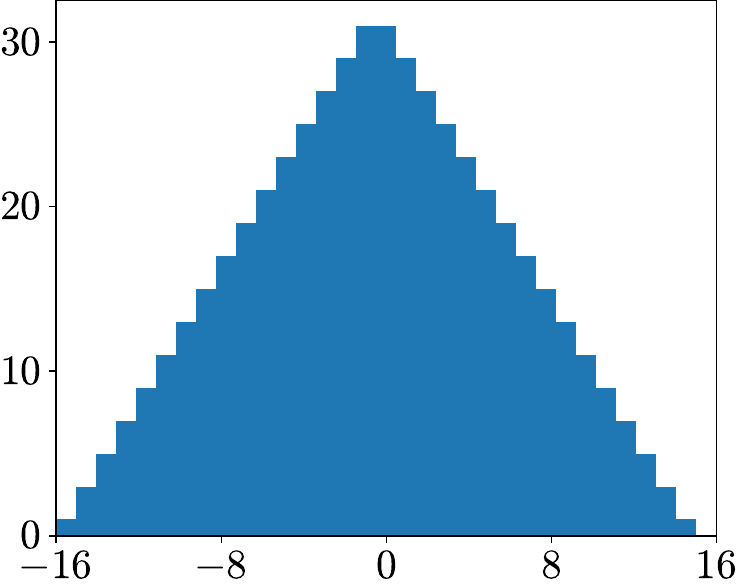} }}%
    \qquad
    \subfloat[\centering 3D]{{\includegraphics[width=0.4\linewidth]{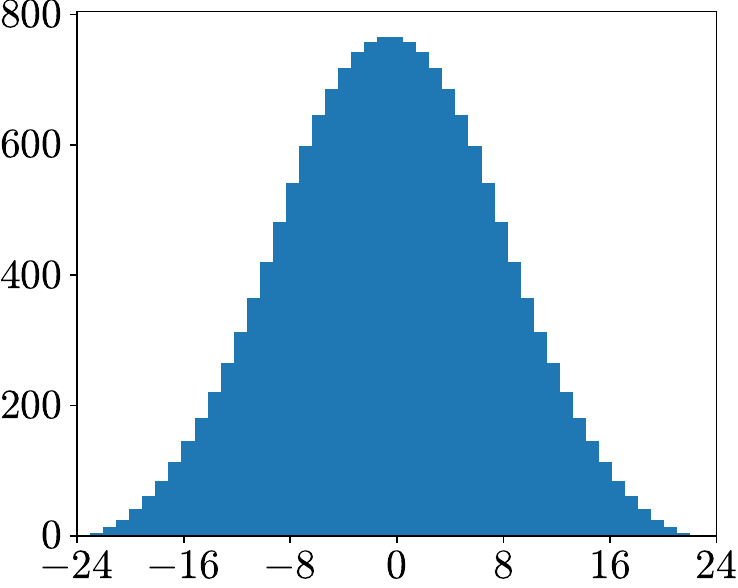} }}%
    \qquad
    \subfloat[\centering 4D]{{\includegraphics[width=0.7\linewidth]{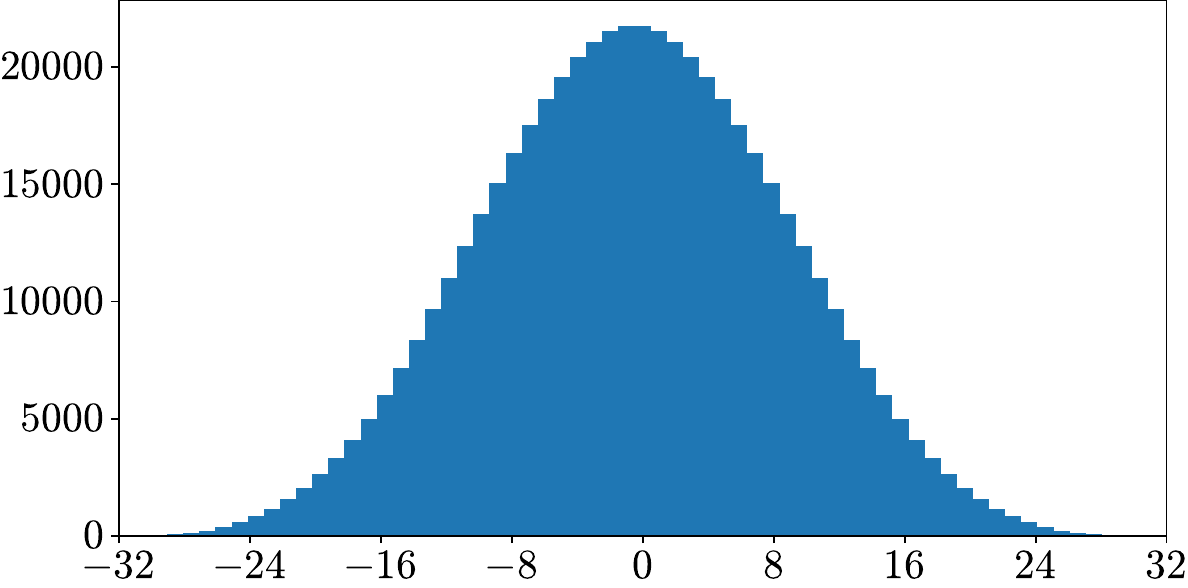} }}%
    \caption{Frequency distributions in $D$-dimensional space where $D{=}2,3,4$. 
    The x-axis is the frequency, and the y-axis denotes the frequency distribution.
    Here, we set $B=8$ and ignore the $2\pi$ term in the frequency. }%
    \label{fig:gaussian_sample}%
\vspace{-0.5em}
\end{figure}

\subsection{RFF in 3D}
\label{app:3d_der}

In the 3D space, we have similar derivations.
The sinusoidal embeddings can be represented as
\begin{align}
    & \phi(u,v,w) \notag\\
    & = \sum_{i=0}^{B-1}\sum_{j=0}^{B-1}\sum_{k=0}^{B-1}
    A_{i,j,k}\mbox{cos}\left(iu\right)\mbox{cos}\left(jv\right)\mbox{cos}\left(kw\right) \notag\\
    & + \sum_{i=0}^{B-1}\sum_{j=0}^{B-1}\sum_{k=0}^{B-1} A_{i,j,k+B}\mbox{cos}\left(iu\right)\mbox{cos}\left(jv\right)\mbox{sin}\left(kw\right) \notag\\
    & + \sum_{i=0}^{B-1}\sum_{j=0}^{B-1}\sum_{k=0}^{B-1} A_{i,j+B,k}\mbox{cos}\left(u\right)\mbox{sin}\left(jv\right)\mbox{cos}\left(kw\right)\notag\\
    & + \sum_{i=0}^{B-1}\sum_{j=0}^{B-1}\sum_{k=0}^{B-1} A_{i,j{+}B,k{+}B}\mbox{cos}\left(iu\right)\mbox{sin}\left(jv\right)\mbox{sin}\left(kw\right) \notag\\
    & + \sum_{i=0}^{B-1}\sum_{j=0}^{B-1}\sum_{k=0}^{B-1} A_{i{+}B,j,k}\mbox{sin}\left(iu\right)\mbox{cos}\left(jv\right)\mbox{cos}\left(kw\right) \notag\\
    & + \sum_{i=0}^{B-1}\sum_{j=0}^{B-1}\sum_{k=0}^{B-1} A_{i{+}B,j,k+B}\mbox{sin}\left(iu\right)\mbox{cos}\left(jv\right)\mbox{sin}\left(kw\right) \notag\\
    & + \sum_{i=0}^{B-1}\sum_{j=0}^{B-1}\sum_{k=0}^{B-1} A_{i{+}B,j+B,k}\mbox{sin}\left(u\right)\mbox{sin}\left(jv\right)\mbox{cos}\left(kw\right) \notag\\
    & + \sum_{i=0}^{B-1}\sum_{j=0}^{B-1}\sum_{k=0}^{B-1} A_{i{+}B,j{+}B,k{+}B}\mbox{sin}\left(iu\right)\mbox{sin}\left(jv\right)\mbox{sin}\left(kw\right)\:, \label{eq:sin_disc_3d}
\end{align}
which needs $K^{3}$ terms for each combination of $i$, $j$ and $k$. 
Here we omit $2\pi$ of each frequency for simplicity.
Following the same trigonometric identities,~\cref{eq:sin_disc_3d} becomes,
\begin{align}
    & \phi(u,v,w) \notag\\
    & = \sum_{i=0}^{B-1}\sum_{j=0}^{B-1} \sum_{k=0}^{B-1} C_{i,j,k}\mbox{cos}\left(2\pi(iu+jv+kw)\right) \notag\\
    & + \sum_{i=0}^{B-1}\sum_{j=0}^{B-1} \sum_{k=0}^{B-1} C_{i,j,k{+}B}\mbox{cos}\left(2\pi(iu+jv-kw)\right) \notag\\
    & + \sum_{i=0}^{B-1}\sum_{j=0}^{B-1} \sum_{k=0}^{B-1} C_{i,j{+}B,k}\mbox{cos}\left(2\pi(iu-jv+kw)\right) \notag\\
    & + \sum_{i=0}^{B-1}\sum_{j=0}^{B-1} \sum_{k=0}^{B-1} C_{i,j{+}B,k{+}B}\mbox{cos}\left(2\pi(iu-jv-kw)\right) \notag\\
    & + \sum_{i=0}^{B-1}\sum_{j=0}^{B-1} \sum_{k=0}^{B-1} C_{i{+}B,j,k}\mbox{sin}\left(2\pi(iu+jv+kw)\right) \notag\\
    & + \sum_{i=0}^{B-1}\sum_{j=0}^{B-1} \sum_{k=0}^{B-1} C_{i{+}B,j,k{+}B}\mbox{sin}\left(2\pi(iu+jv-kw)\right) \notag\\
    & + \sum_{i=0}^{B-1}\sum_{j=0}^{B-1} \sum_{k=0}^{B-1} C_{i{+}B,j{+}B,k}\mbox{sin}\left(2\pi(iu-jv+kw)\right) \notag\\
    & + \sum_{i=0}^{B-1}\sum_{j=0}^{B-1} \sum_{k=0}^{B-1} C_{i{+}B,j{+}B,k{+}B}\mbox{sin}\left(2\pi(iu-jv-kw)\right)\:. \label{eq:sin_disc_3d_t}
\end{align}

Similarly to the 2D case, these $K^3$ samples of frequencies can also be sub-sampled according to the shape of frequency distributions as in~\cref{fig:gaussian_sample} (b). 
The sinusoidal embeddings are expected to work more efficiently in higher dimensions and this shape becomes more and more similar to a Gaussian distribution (\cref{fig:gaussian_sample} (c)), which leads to the definition of RFF embedding.

\subsection{Frequency distribution derivation}
\label{app:fd_der}
As the dimension~$D$ of the points increases, the efficiency of these band-limited embeddings becomes more prominent. 
The shape of the frequency distribution becomes the sum of $D$ uniform distributions on $[-B,B]$, which yields the Irwin-Hall distribution. 
According to the Central Limit Theorem, we know that the sum of $n$ uniform distributions on $[0,1]$ (Irwin-Hall distribution) approximate Gaussian distribution with $\mu\:{=}\:\frac{n}{2}$ and $\sigma^{2}\:{=}\:\frac{n}{12}$. 
Therefore, in $D$-dimensional space, the distribution of projection frequencies $\W$ is
\begin{equation}
    \label{eq:CLT}
    \begin{aligned}
        \W&=\sum_{i=1}^{D}\mathcal{U}(-B,B)
        \\&=\sum_{i=1}^{D}(2B\mathcal{U}(0,1)-B)
        \\&= 2B\sum_{i=1}^{D}\mathcal{U}(0,1)-DB
        \\&\approx 2B\mathcal{N}(\frac{D}{2},\frac{D}{12})-DB
        \\&= \mathcal{N}(0,\frac{DB^2}{3}),
    \end{aligned}
\end{equation}
where $\mathcal{U}(a,b)$ is uniform distribution on $[a,b]$ and $\mathcal{N}(\mu,\sigma^2)$ is Gaussian distribution with mean $\mu$ and standard deviation $\sigma$, $D$ is the dimension and $B$ is the band-width.
We therefore know that the projection frequencies $\W$ (as well as the weight matrix in RFF) should follow the Gaussian distribution $\mathcal{N}(0,\frac{DB^2}{3})$.
This gives some intuitions on why RFF is a more attractive embedding for higher dimensional data. 
To achieve comparable performance with $\mbox{sinc}$ or a similar approximated function (\eg, Gaussian), we need $K^{D}$ samples. 
In contrast, for RFF, only a limited amount of sub-samples is needed. 
Additionally, when tuning the standard deviation of the scale of RFF (standard deviation of the Gaussian distribution), one can easily control the bandwidth of this encoder.

\section{Pooling functions}
\label{sec:pool}
Thus far, the encoding of the point cloud discussed in the main text and aforementioned section all used sum pooling to get the final point representation.
However, max pooling is the most common pooling mechanism used in the context of learned PPEs such as PointNet~\cite{qi2017pointnet}.
Mean pooling is also used as a comparison oftentimes. 
Here, we first show that all these pooling methods can be formulated as a form of sum pooling. 
Then we dive deeper into the difference between max pooling and mean pooling.

Mean pooling is straightforwardly connected to sum pooling.
The only difference between these two pooling functions is a scale parameter:
\begin{equation}
    \label{eq:meantosum}
    \begin{aligned}
        \y & = \mbox{mean}\left(\{h(\p_1),h(\p_2),{\cdots},h(\p_N)\}\right) \\ 
        & = \frac{1}{N}\sum_{i=1}^{N}h(\p_i) \\
        & = \sum_{i=1}^{N}\frac{1}{N}h(\p_i) \\
        & = \sum_{i=1}^{N}h'(\p_i)\:,
    \end{aligned}
\end{equation}
where $h'(\cdot)\:{=}\:\frac{1}{N}h(\cdot)$. 
In~\cref{eq:meantosum}, we take the scale parameter into the PPE $h$ and convert mean pooling to sum pooling. 
Max pooling can be converted similarly since max pooling can be seen as zero masking all the elements in one column except for the maximum value in that column and summing over columns. 
If we take the zero masking process into $h'(\cdot)$, then we have
\begin{equation}
    \label{eq:maxtosum}
        h'(\p_i)[k]=
        \begin{cases}
        h(\p_i)[k],&\:\text{if }\forall j\neq i,h(\p_i)[k]\ge h(\p_j)[k]\:, \\
        0,&\:\text{otherwise}\:.
        \end{cases}
\end{equation}
Note that we ignore the case when there are multiple maximum values in one column, which rarely happens in real-world data.
Following this analogy, any pooling functions can be converted to sum pooling. 
The magic step is to take all the operations before the last sum operation into the PPE function. 
Therefore, the bandwidth and smoothness theory we discussed in the main text and aforementioned sections holds for all types of point encoders.

In PointNet~\cite{qi2017pointnet} and its variants~\cite{qi2017pointnet++, wang2019dynamic, wu2019pointconv, liu2019relation}, max pooling has been the most popular pooling choice due to its good performance over mean pooling. 
However, the reason why max pooling has better performance is unclear. 
In order to have a better understanding of the difference between max pooling and mean pooling, we show an extreme case where the PPE only contains one linear layer,~\ie, $h(\x) \:{=}\: \A\x{+}\b$, without $\mbox{ReLU}$ or any other nonlinear activation functions. 
In this case, mean pooling fails while max pooling works surprisingly well. 
With mean pooling, the entire network can be represented as
\begin{equation}
    \label{equ:linear_mean}
    \y = \frac{1}{N}\sum_{i=1}^N(\A\x_i+\b)=\A\Bar{\x}+\b\:,
\end{equation}
where $\Bar{\x}\:{=}\:\frac{1}{N}\sum_{i}^{N}\x_{i}$ is the center (mean) of the point cloud. 
In this case, mean pooling fails since the feature is only a linear transformation of the center point. 
This can also be explained by the bandwidth and smoothness properties. 
Even though the distance information is perfectly preserved, the bandwidth of the PPE is always $3$ (for 3D point cloud) when using mean pooling, which does not increase at all.
However, max pooling acts as a nonlinear operator.
\cref{eq:maxtosum} shows that the max pooling operator is similar to an extreme ReLU function. 
Following this observation, we introduce the following lemma.
\begin{lem}
    For an input $\X\:{\in}\:\mathbb{R}^{N {\times} K}$, max pooling by column is equivalent to a sequence of max-normalization (parameterized by $\b$ and $c$, which are dependent on $\X$, similar to a weaker version of Instance Normorlizaiton), ReLU and a sum pooling by column, which can be written as
    \begin{equation}
        \label{seq:max_mean}
        \max\X= \mathrm{mean}(\mathrm{ReLU}(\frac{\X-\b}{c}))
    \end{equation}
\end{lem}
As discussed in~\cite{zheng2022trading}, ReLU can inherently increase bandwidth and keep some distance at the same time, so max pooling can work well even with only a linear layer.

\subsection{Proof of Lemma 1}
Now let us consider how we do max pooling given a matrix $\X\:{\in}\:\mathbb{R}^{N {\times} K}$.  
We want to apply max pooling to its columns to get a feature vector $\hat{\x}\:{\in}\: \mathbb{R}^{1 {\times} K}$, which is equivalent to first zero masking all the elements except for the maximum values in each column as $\ddot{\X}$), and then do a sum pooling
\begin{equation}
\label{eq:max_sum}
\begin{aligned}
    \hat{\x} &= \max (\X) \\
    &= \begin{bmatrix} \max (\X_{:,1}) & \max (\X_{:,2}) & \cdots & \max (\X_{:,K}) \end{bmatrix} \\
    &= \begin{bmatrix} \mbox{sum} (\ddot{\X}_{:,1}) & \mbox{sum} (\ddot{\X}_{:,2}) & \cdots & \mbox{sum} (\ddot{\X}_{:,K}) \end{bmatrix} \\
    &= \mbox{sum} (\ddot{\X}) \:.
\end{aligned}
\end{equation}
The maximum value masking operation is similar to a ReLU function.
If we represent $\ddot{\X}$ using a ReLU function, we can obtain the masked matrix by setting all elements except the maximum value in each column to be negative.
Specifically, we can achieve this by subtracting a value $\tilde{x}_i$ from each element in the column, where $\tilde{x}_i$ is between the largest value $m_i$ and the second-largest value $n_i$ in column $i$, for $i\:{=}\:1,\dots,K$.
After the ReLU operation, since all the elements except the largest value are zeros in each column, we can multiply each column of the matrix by $\tilde{c}_i \:{=}\: \frac{m_i}{m_i-\tilde{x}_i}$ to get the maximum value back:
\begin{equation}
    \hat{\ddot{\X}} = \tilde{\mathbf{c}} \circ \mbox{ReLU}( \X - \tilde{\x}) \:,
\end{equation}
where $\tilde{\mathbf{c}} \:{=}\: [\frac{m_1}{m_1 \:{-}\: \tilde{x}_1} \frac{m_2}{m_2 \:{-}\: \tilde{x}_2},{\cdots},\frac{m_K}{m_K \:{-}\: \tilde{x}_K}]$ is the multiple for each column, and $\tilde{\x} \:{=}\: [\tilde{x}_1,\tilde{x}_2,{\cdots},\tilde{x}_K]$ is deducted value in each column. 
It is worth noting that the choice of $\tilde{x}_i$ is flexible and can be any value between $m_i$ and $n_i$, which makes $\tilde{c}_i \:{=}\: \frac{m_i}{m_i \:{-}\: \tilde{x}_i}$ vary. 
However, we can choose $\tilde{x}_i$ such that all the $\tilde{c}_i$ take the same value, say $\frac{1}{\tilde{c}}$, and feed this common value to the ReLU function to get
\begin{equation}
    \hat{\ddot{\X}} = \mbox{ReLU}\left( \frac{\X - \tilde{\x}}{\tilde{c} }\right) \:.
\end{equation}
The part inside the ReLU function can be viewed as a form of batch normalization. 
In total, the operation by column is equivalent to a sequence of max-normalization, which is parameterized by $\tilde{\x}$ and $\tilde{c}$, both of which depend on the input matrix $\X$. 
The max pooling therefore can be written as a ReLU function and a column-wise mean pooling function:
\begin{equation}
\label{eq:max_mean}
    \hat{\x} = \max (\X) = \mbox{mean}\left(\mbox{ReLU}\left(\frac{\X-\tilde{\x}}{\tilde{c}}\right)\right) \:.
\end{equation}
Compared to mean pooling, max pooling contains a weak normalization. 
As a result, if we add a max pooling to a linear PPE, then the entire network can be written as
\begin{equation}
    \label{equ:linear_max}
    \y = \max (\A\x_i+\b) = \mbox{mean}\left( \mbox{ReLU}\left(\frac{\A\x_i+\b-\tilde{\x}}{\tilde{c}}\right)\right) \:,
\end{equation}
where $\tilde{\x}$ and $\tilde{c}$ are parameters that depend on $\x_i$. 
Therefore, max pooling works well even with a single-layer linear PPE.
Specifically, it is equivalent to mean pooling with a linear normalization and ReLU function.
After carefully investigating this simple example, we prove that max pooling has inherent nonlinearity that leads to better performance than mean pooling in MLPs.
Even when we use multiple layers with ReLU functions, this nonlinearity helps increase the bandwidth of PPE, the performance is still slightly better than using mean pooling. 
However, max pooling is only applicable for an MLP PPE specifically in PointNet. 
If a Positional Encoder is used, the PE function itself already contains nonlinearity and the bandwidth is already high enough.
As a result, the advantage of max pooling is meaningless in PE-based PPEs.

\section{Experiments}
\label{sec:supp:exp}

\subsection{Results of other data corruptions}
\begin{figure*}[t!]
    \centering
    \subfloat[\centering]{{\includegraphics[width=0.45\linewidth]{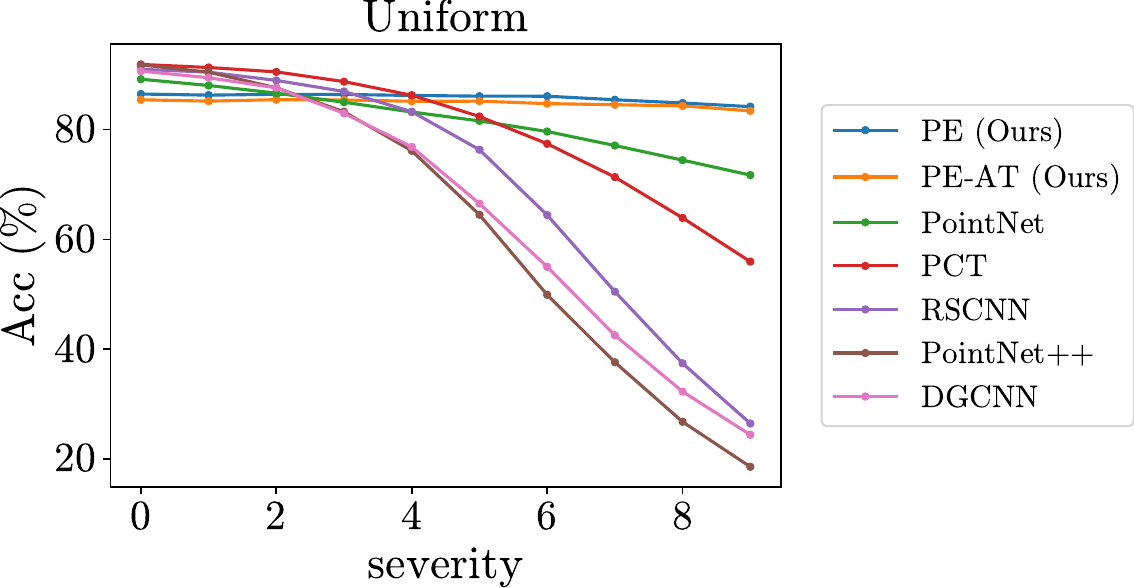} }}%
    \qquad
    \subfloat[\centering]{{\includegraphics[width=0.45\linewidth]{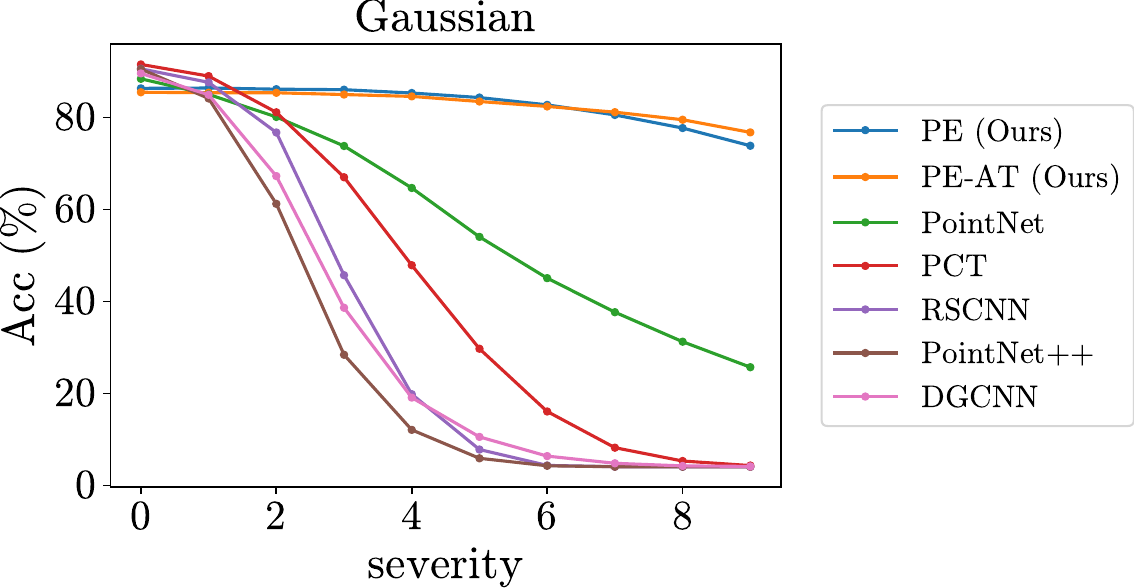} }}%
    \qquad
    \subfloat[\centering]{{\includegraphics[width=0.45\linewidth]{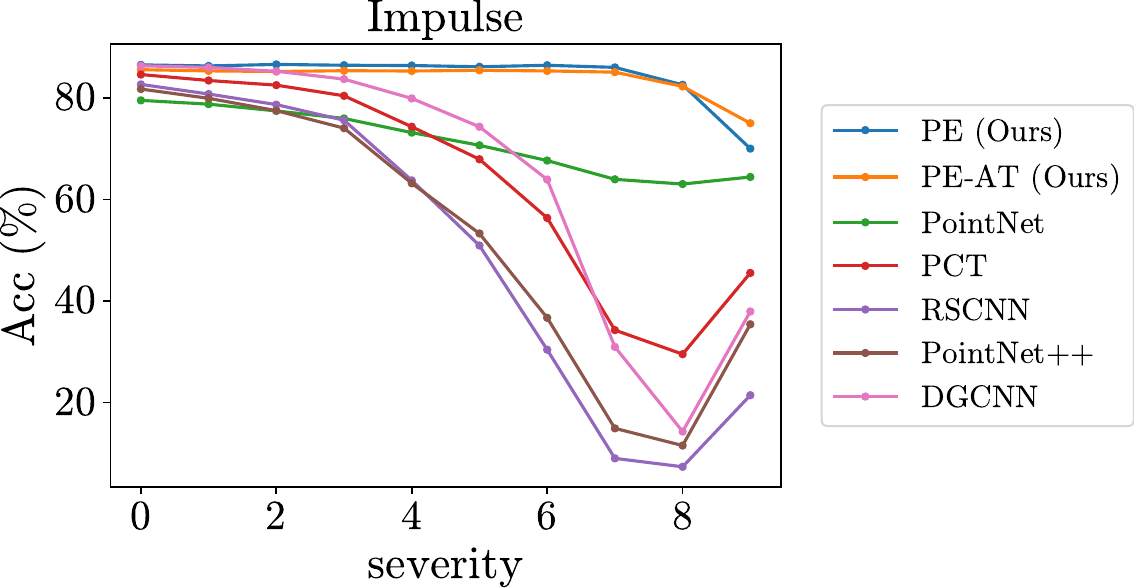} }}%
    \qquad
    \subfloat[\centering]{{\includegraphics[width=0.45\linewidth]{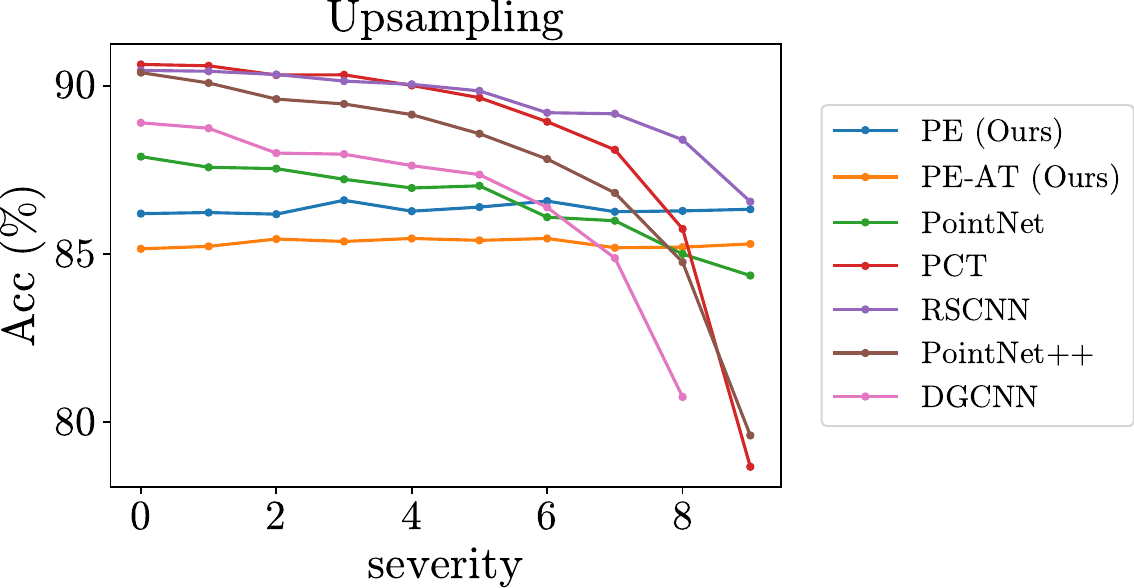} }}%
    \qquad
    \subfloat[\centering]{{\includegraphics[width=0.45\linewidth]{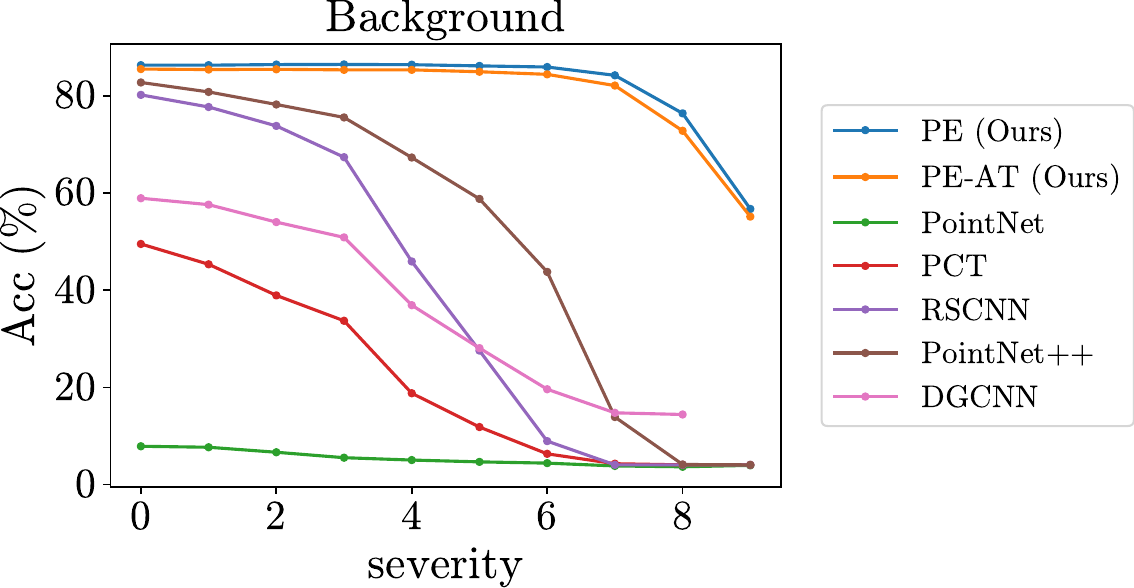} }}%
    \qquad
    \subfloat[\centering]{{\includegraphics[width=0.45\linewidth]{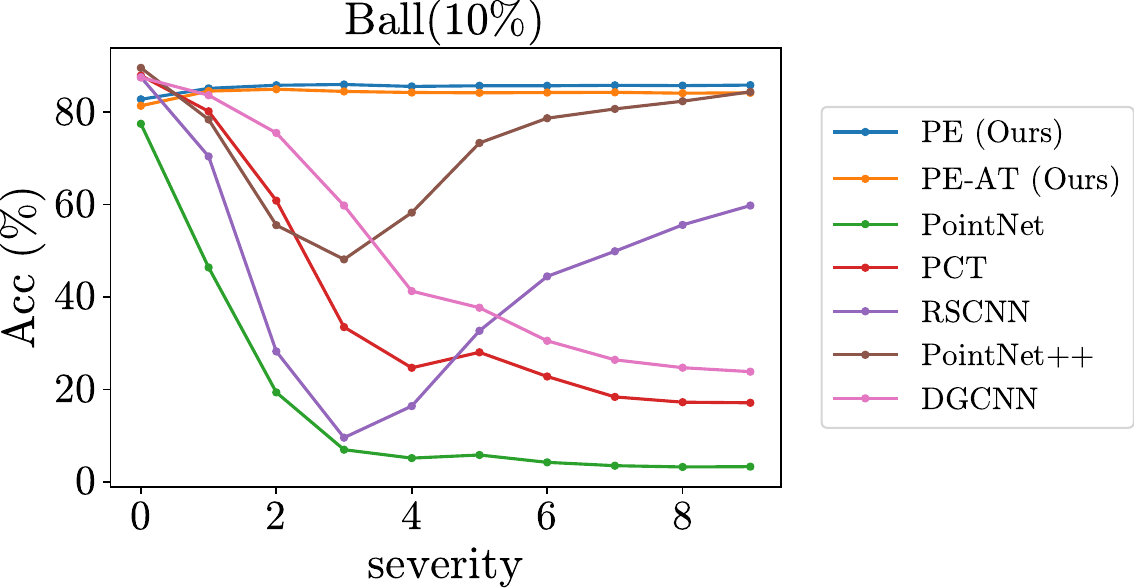} }}%
    \qquad
    \subfloat[\centering]{{\includegraphics[width=0.45\linewidth]{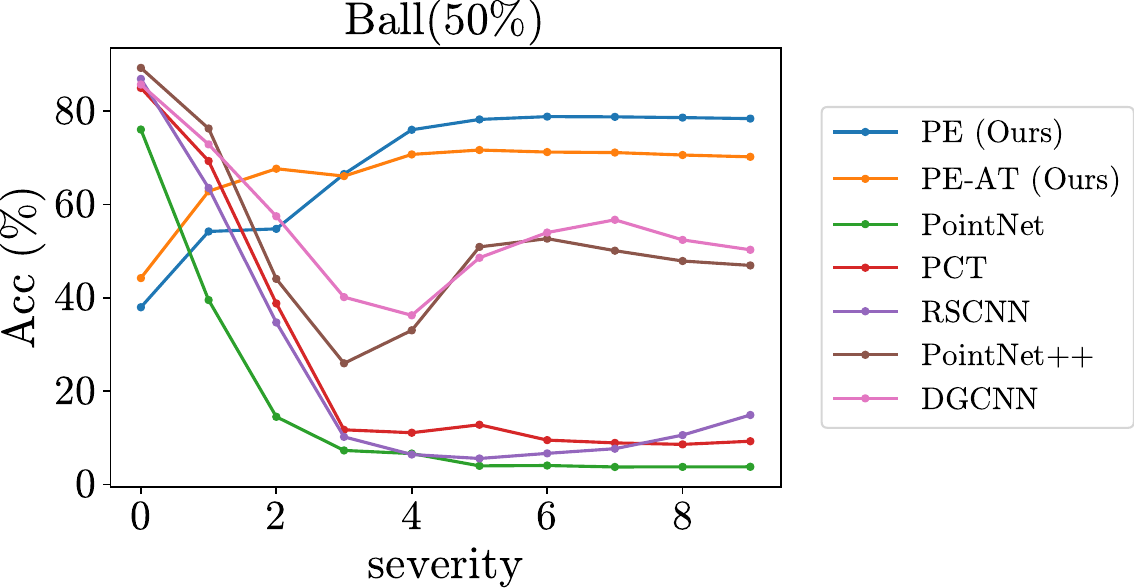} }}%
    \qquad
    \subfloat[\centering]{{\includegraphics[width=0.45\linewidth]{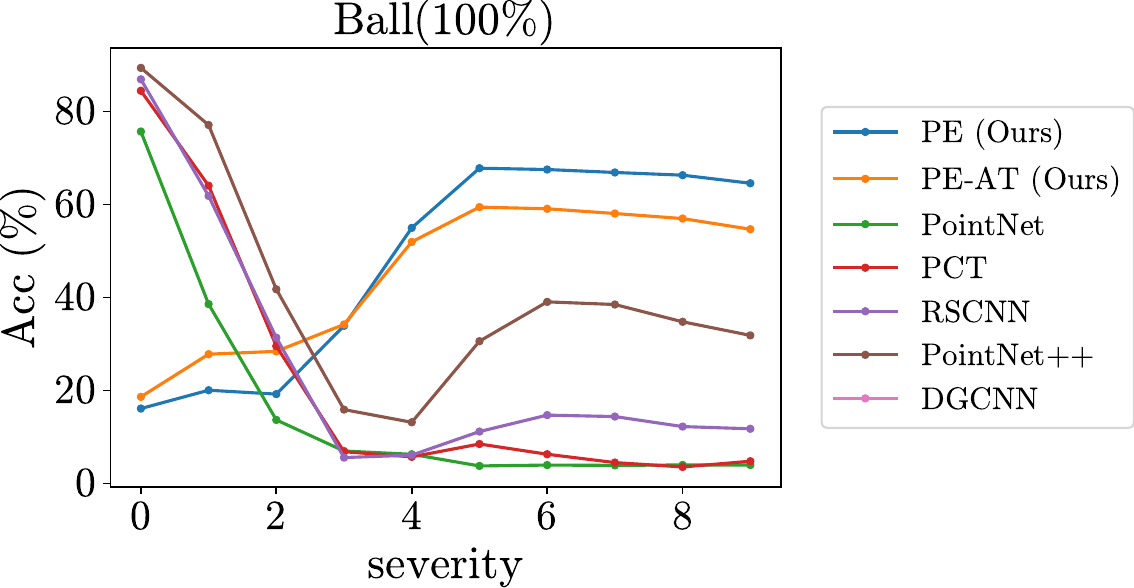} }}%
    \caption{Classification accuracy on our data corruption dataset. 
    As severity increases, corruption becomes stronger. 
    However, the meaning of severity varies between corruption types, such as the variance or range of the noise and the number of outliers added. 
    DGCNN fails to get reasonable performance when the number of input data is too large.
    Our proposed PE-based PPE methods maintain robust and good performance across almost all category of data corruptions.}%
    \label{fig:cor_acc}%
\end{figure*}

\begin{figure*}[t!]
    \centering
    \subfloat[\centering]{{\includegraphics[width=0.45\linewidth]{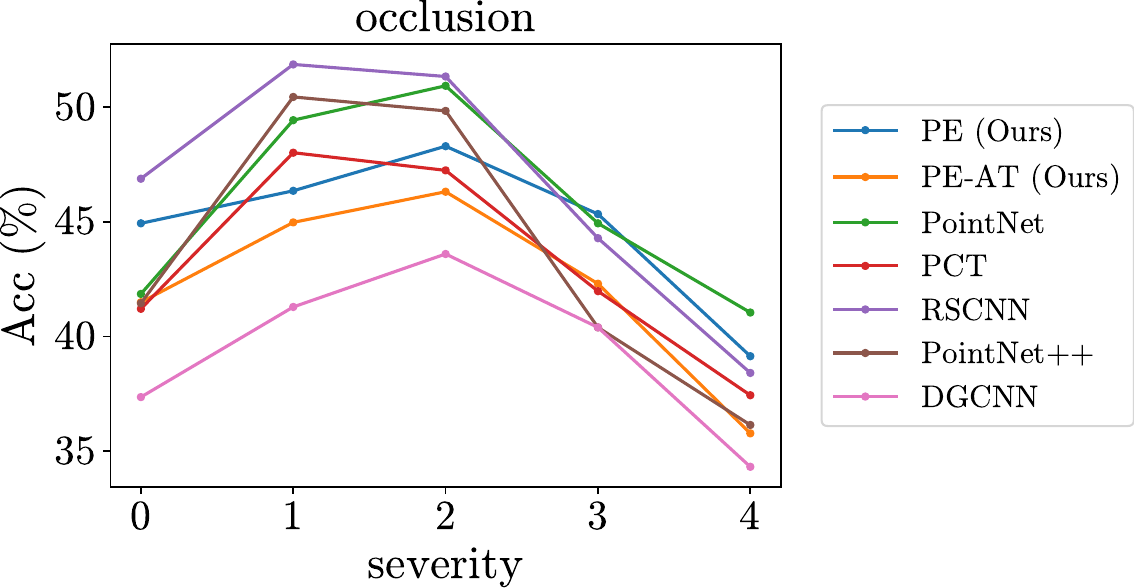} }}%
    \qquad
    \subfloat[\centering]{{\includegraphics[width=0.45\linewidth]{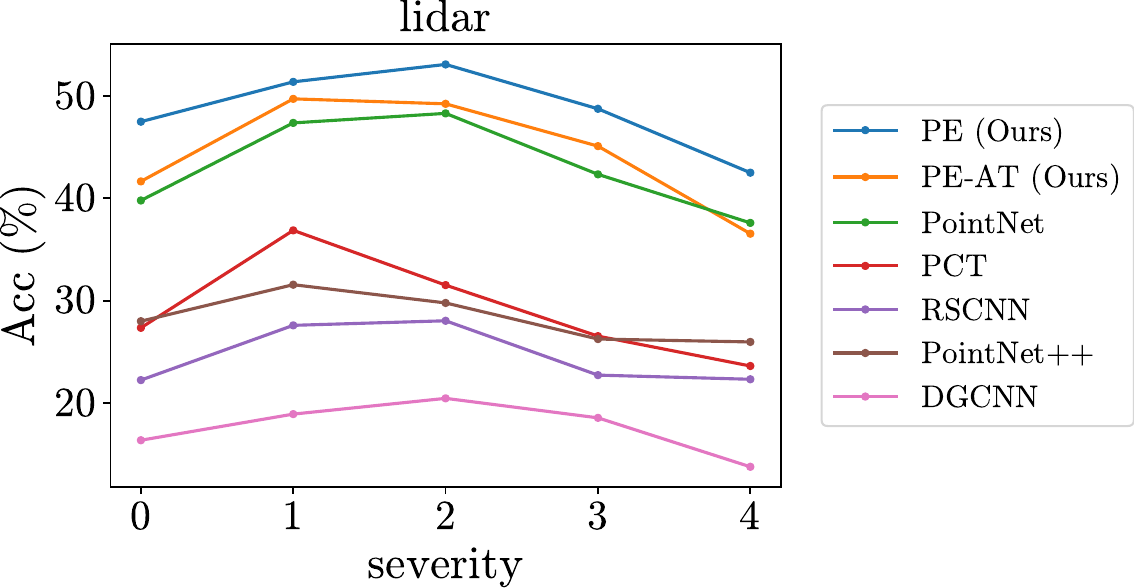} }}%
    \qquad
    \subfloat[\centering]{{\includegraphics[width=0.45\linewidth]{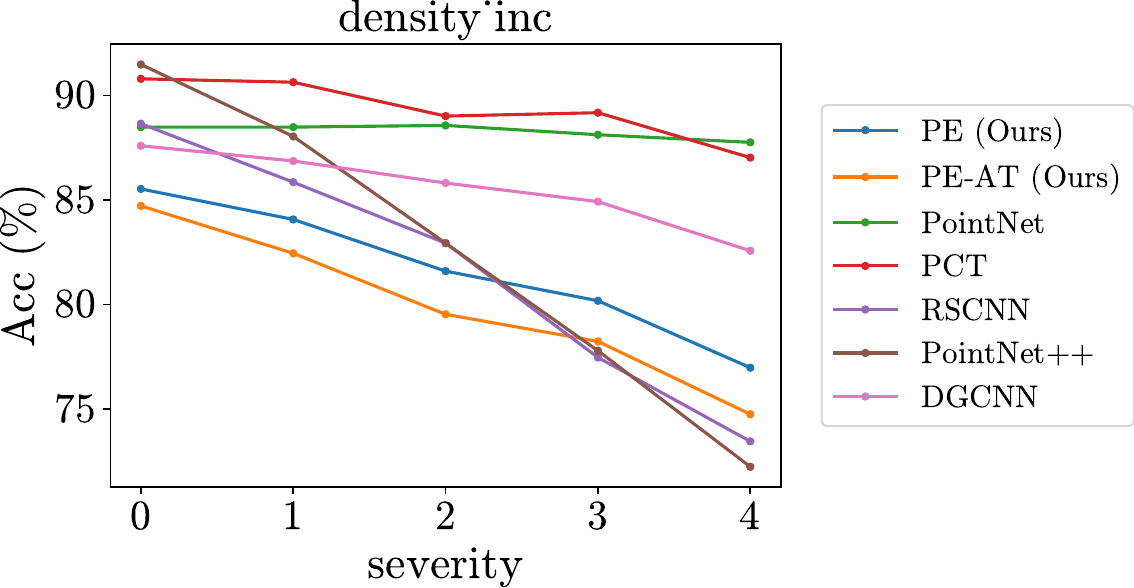} }}%
    \qquad
    \subfloat[\centering]{{\includegraphics[width=0.45\linewidth]{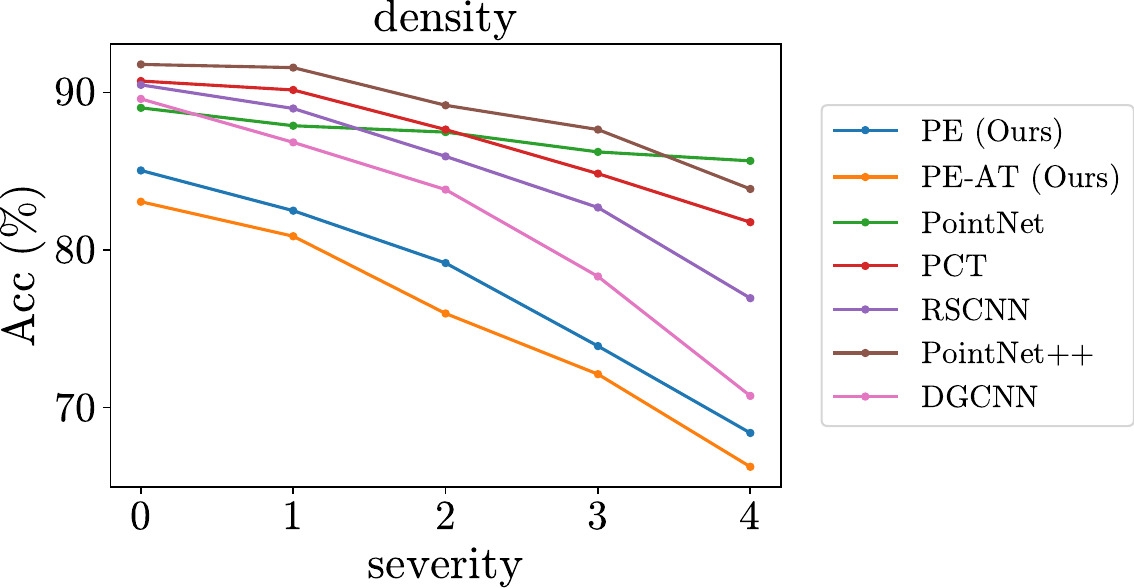} }}%
    \qquad
    \subfloat[\centering]{{\includegraphics[width=0.45\linewidth]{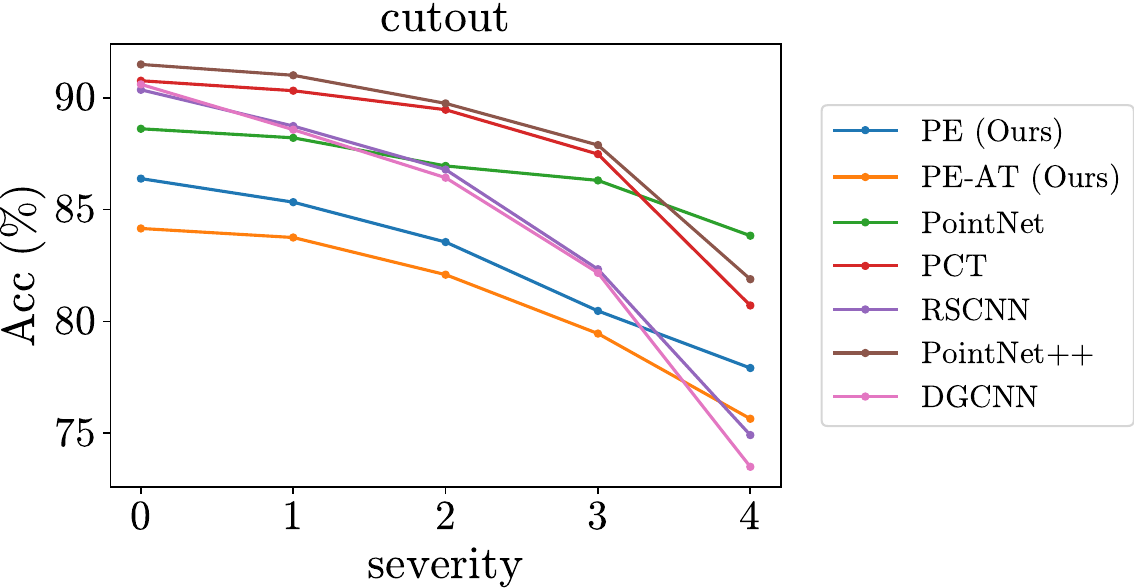} }}%
    \qquad
    \subfloat[\centering]{{\includegraphics[width=0.45\linewidth]{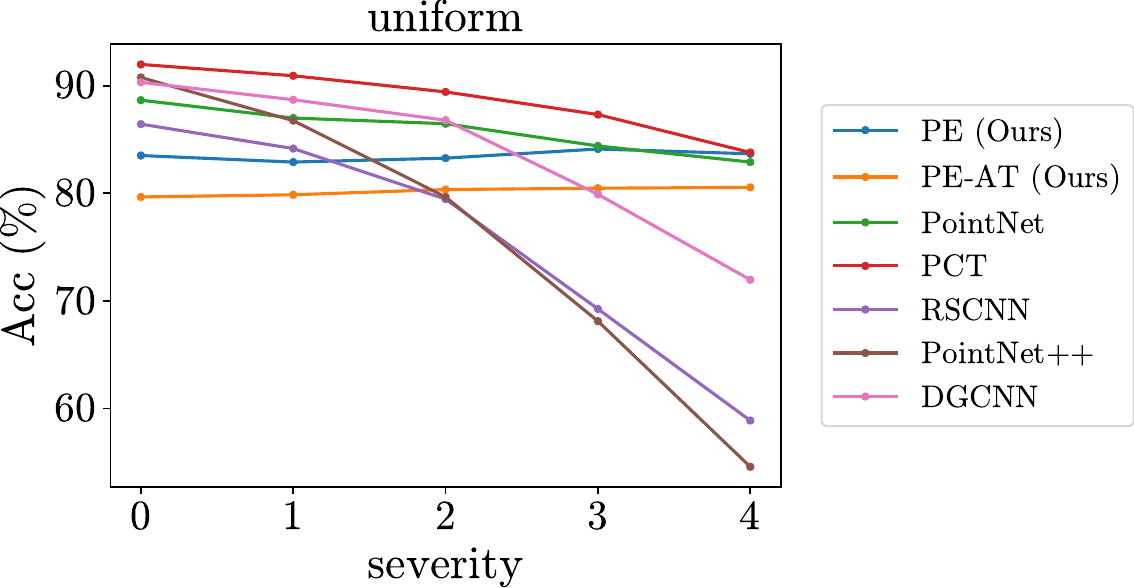} }}%
    \qquad
    \subfloat[\centering]{{\includegraphics[width=0.45\linewidth]{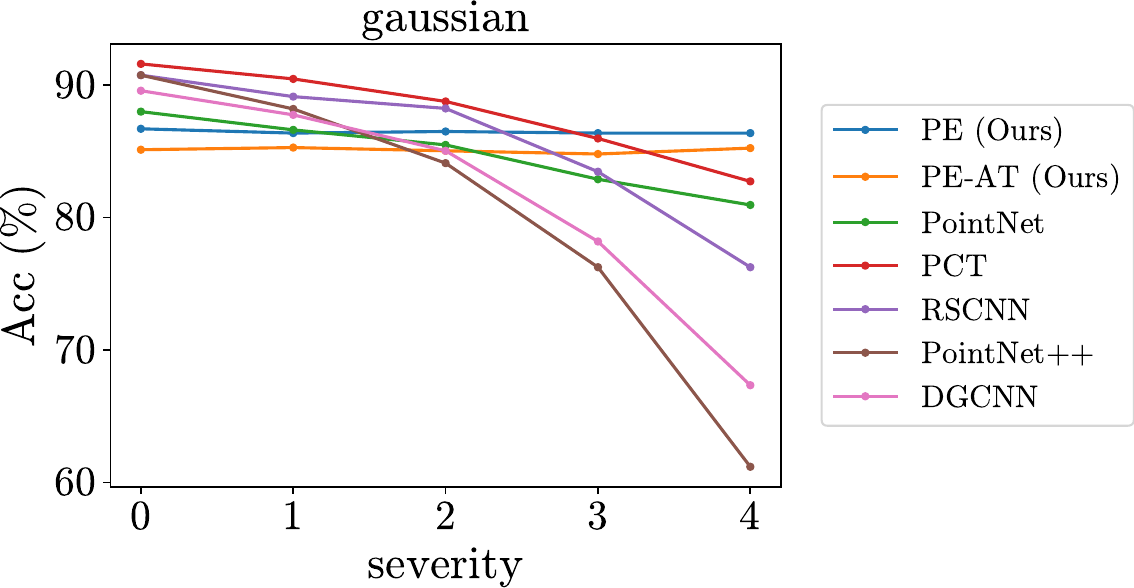} }}%
    \qquad
    \subfloat[\centering]{{\includegraphics[width=0.45\linewidth]{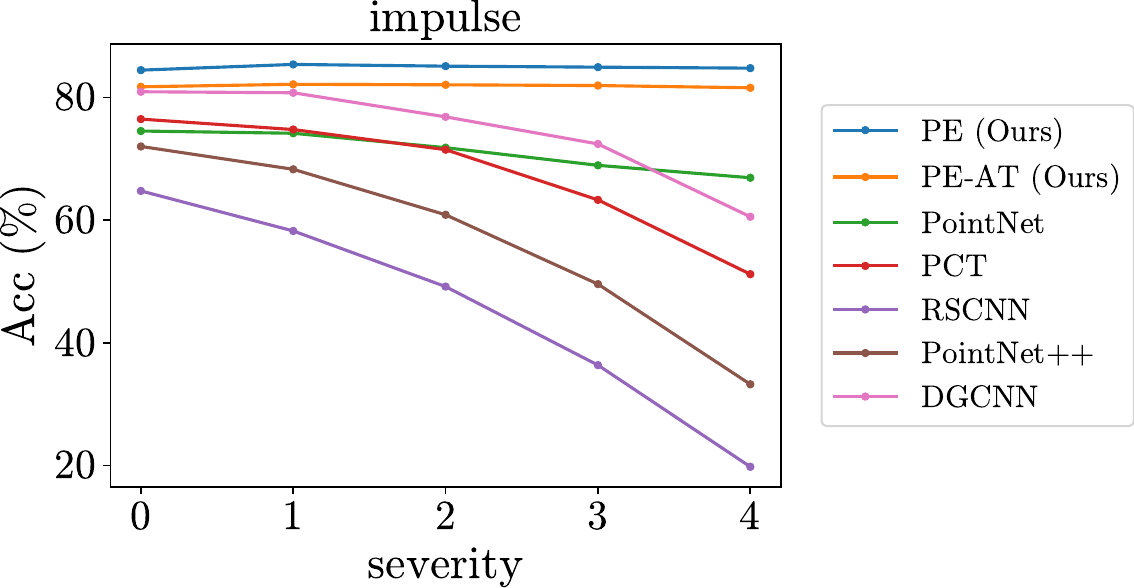} }}%
    \caption{Classification accuracy on the original ModelNet40-C~\cite{sun2022benchmarking} dataset. Continued in~\cref{fig:modelnetc_acc_2}}%
    \label{fig:modelnetc_acc_1}%
\end{figure*}

\begin{figure*}[t!]
    \centering
    \subfloat[\centering]{{\includegraphics[width=0.45\linewidth]{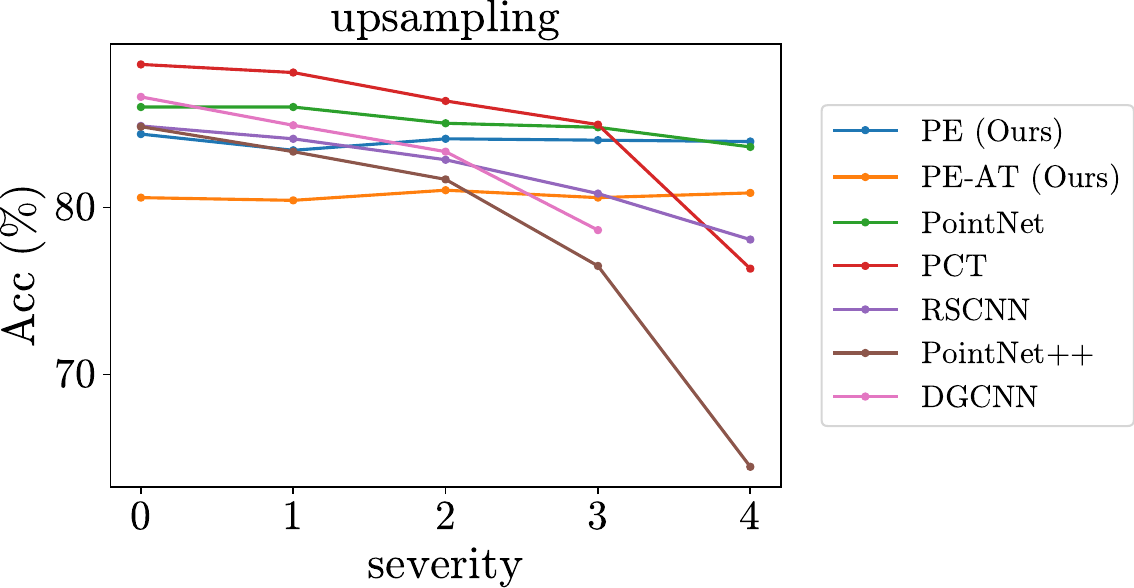} }}%
    \qquad
    \subfloat[\centering]{{\includegraphics[width=0.45\linewidth]{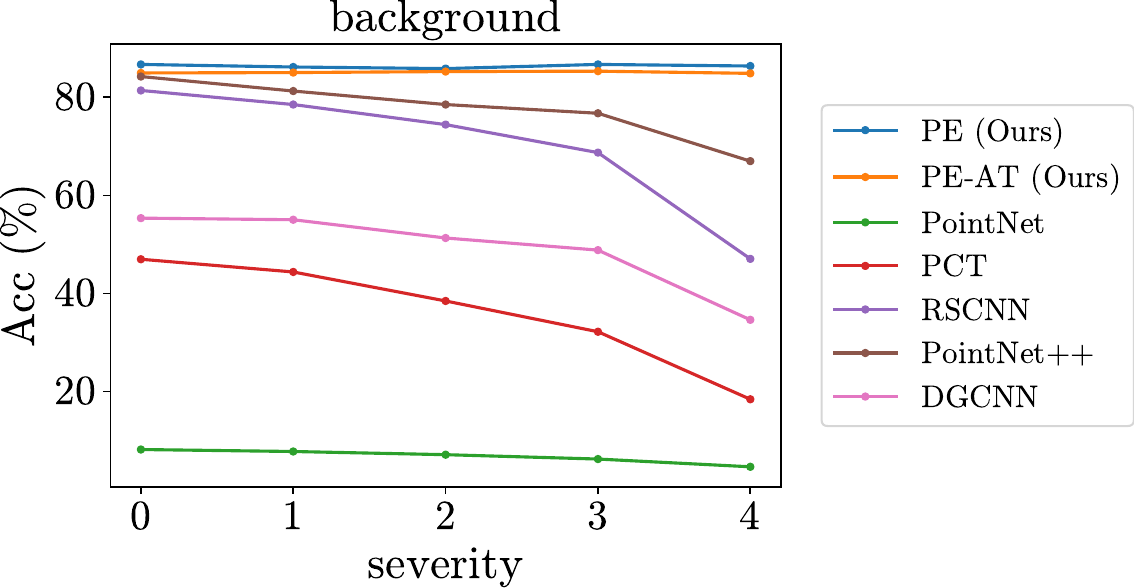} }}%
    \qquad
    \subfloat[\centering]{{\includegraphics[width=0.45\linewidth]{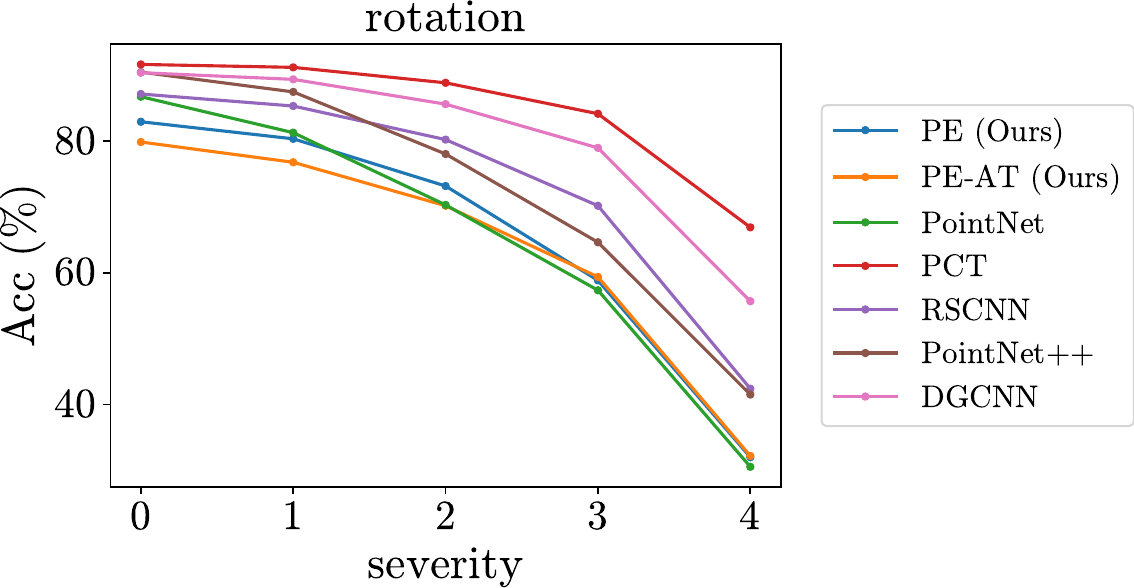} }}%
    \qquad
    \subfloat[\centering]{{\includegraphics[width=0.45\linewidth]{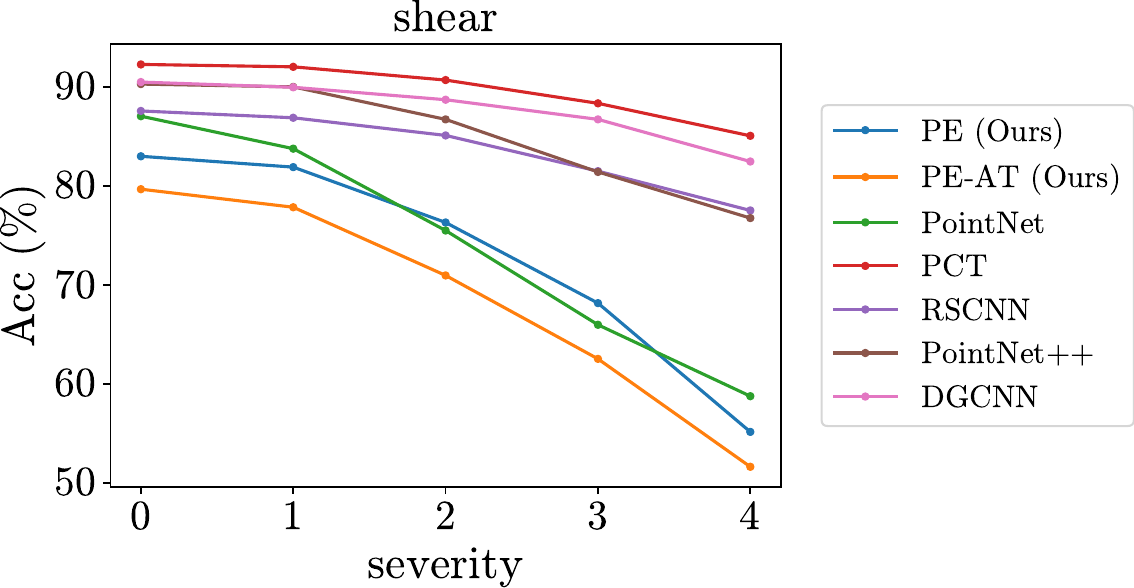} }}%
    \qquad
    \subfloat[\centering]{{\includegraphics[width=0.45\linewidth]{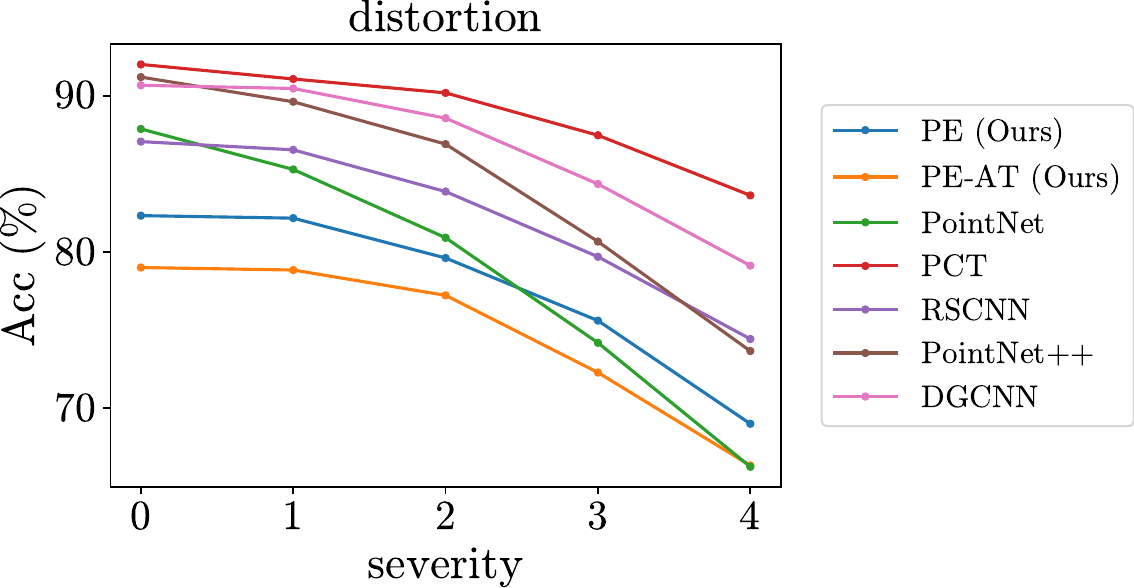} }}%
    \qquad
    \subfloat[\centering]{{\includegraphics[width=0.45\linewidth]{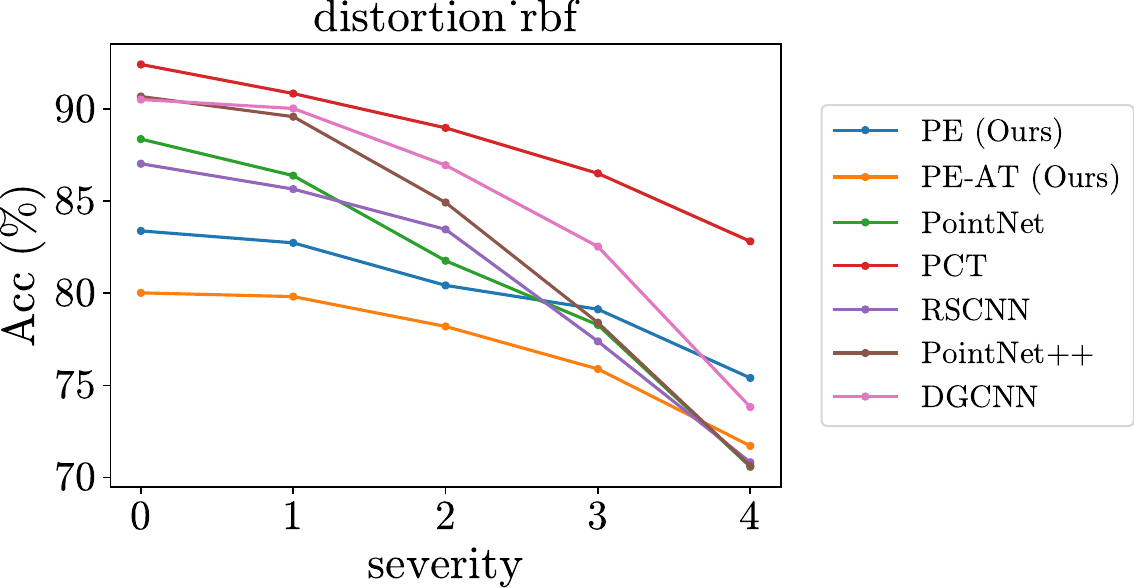} }}%
    \qquad
    \subfloat[\centering]{{\includegraphics[width=0.45\linewidth]{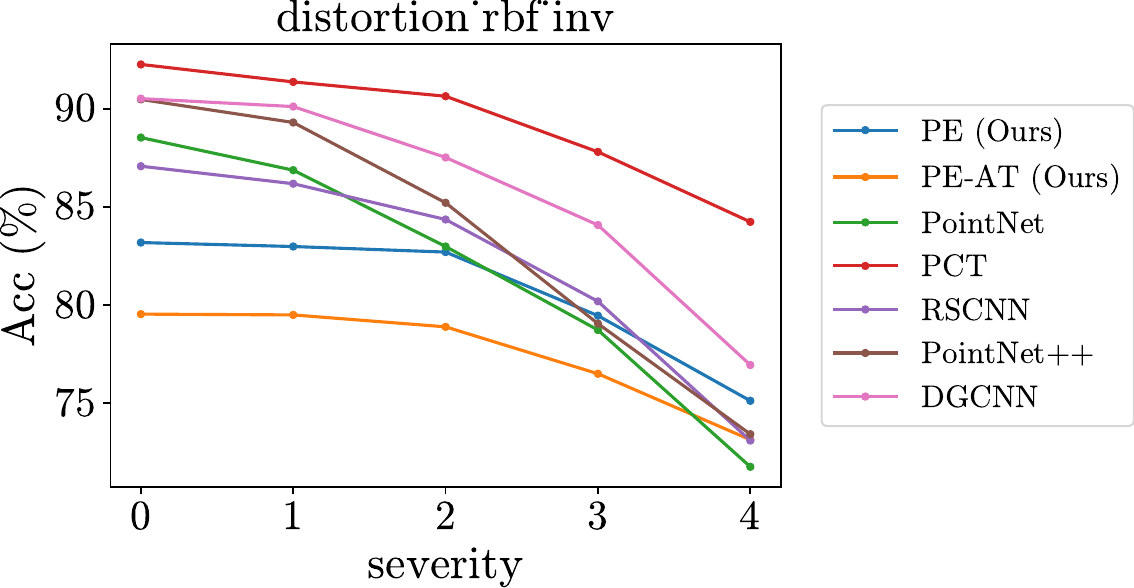} }}%
    \caption{Classification accuracy on the original ModelNet40-C~\cite{sun2022benchmarking} dataset. 
    As severity increases, corruption becomes stronger. 
    Note that the severity is smaller than our modified dataset, which is less challenging. 
    In this case, the clean part of the points (\eg, the original points) may dominate the entire point cloud, yielding good performance for learned PPEs.}
    \label{fig:modelnetc_acc_2}%
\end{figure*}

We provide the full results of different models tested under different severity of different data corruptions on our modified dataset in~\cref{fig:cor_acc}. 
The full results of the original ModelNet-C~\cite{sun2022benchmarking} dataset are shown in~\cref{fig:modelnetc_acc_1} and~\cref{fig:modelnetc_acc_2}.

The quantitative classification results of different models tested under data corruptions other than in the main text are shown in~\cref{tab:modelnetc}.

\begin{figure*}[t]
    \centering
    \subfloat[\centering Uniform]{{\includegraphics[width=0.3\linewidth]{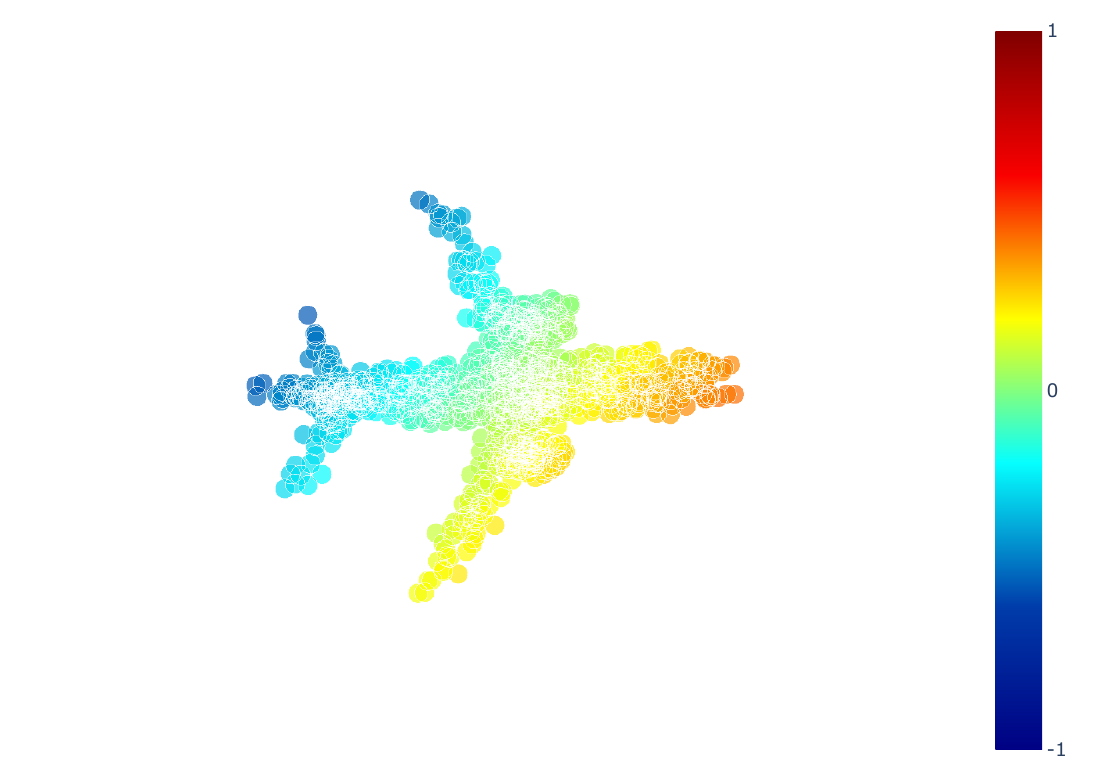} }}%
    \qquad
    \subfloat[\centering Gaussian]{{\includegraphics[width=0.3\linewidth]{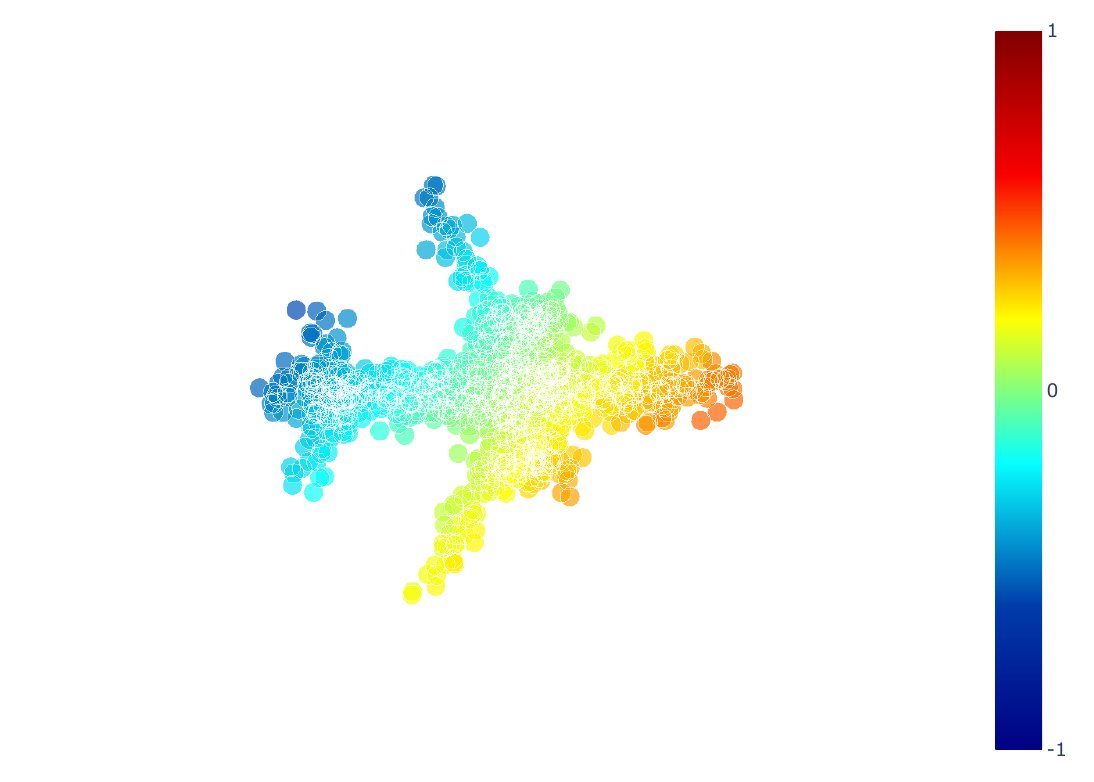} }}%
    \qquad
    \subfloat[\centering Impulse]{{\includegraphics[width=0.3\linewidth]{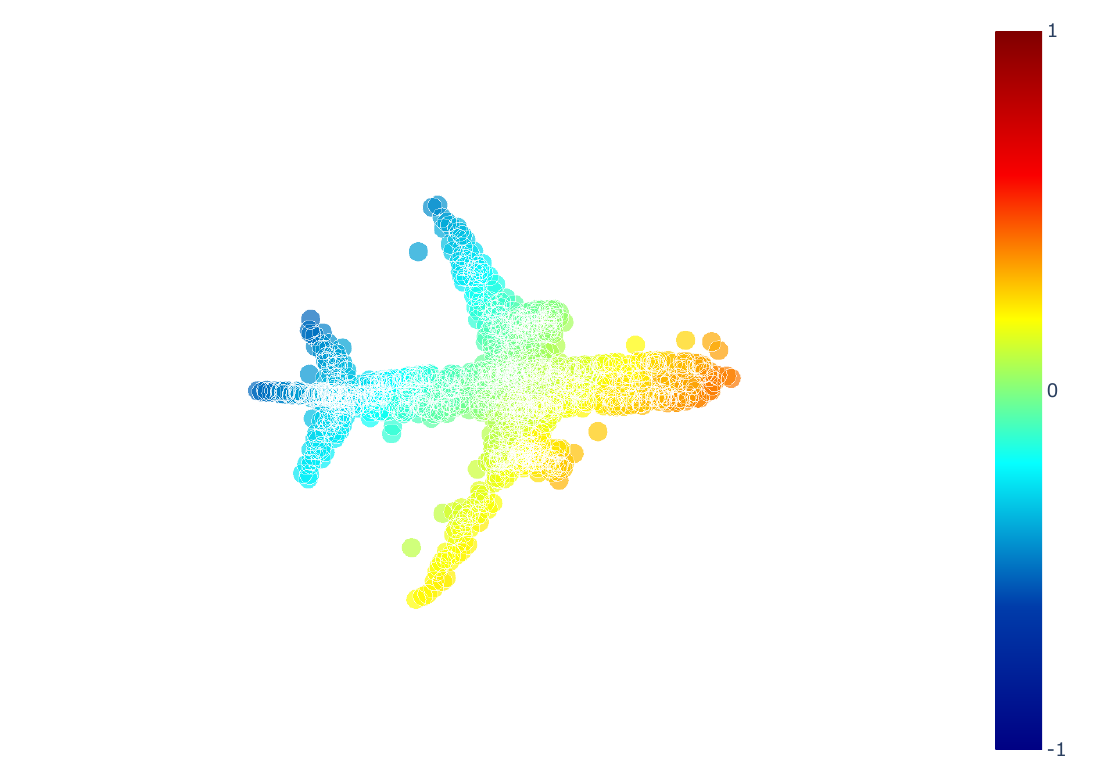} }}%
    \qquad
    \subfloat[\centering Upsampling]{{\includegraphics[width=0.3\linewidth]{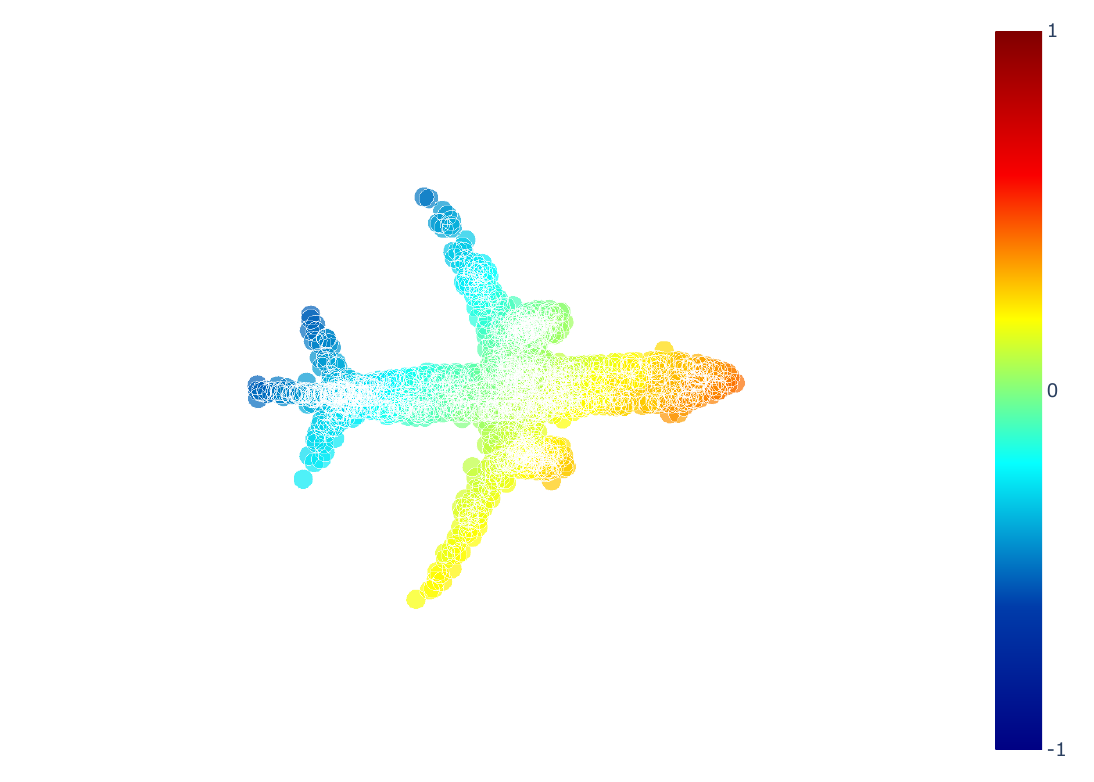} }}%
    \qquad
    \subfloat[\centering Background]{{\includegraphics[width=0.3\linewidth]{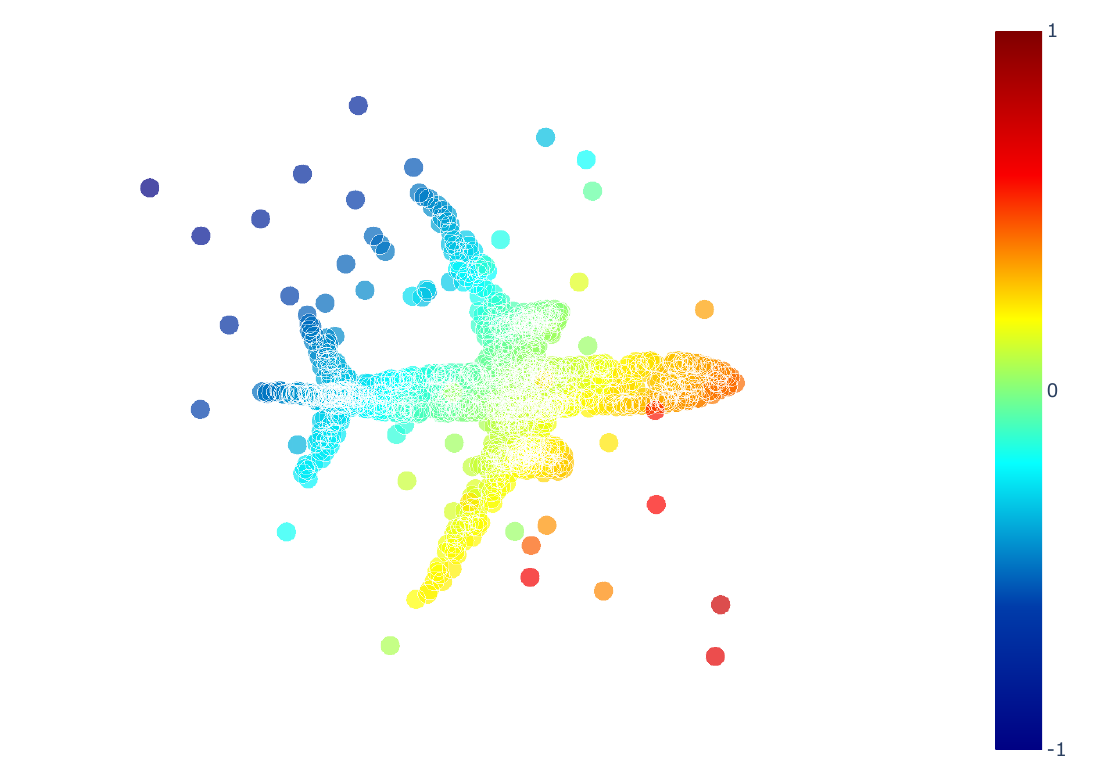} }}%
    \qquad
    \qquad
    \subfloat[\centering Ball $(10\%)$]{{\includegraphics[width=0.3\linewidth]{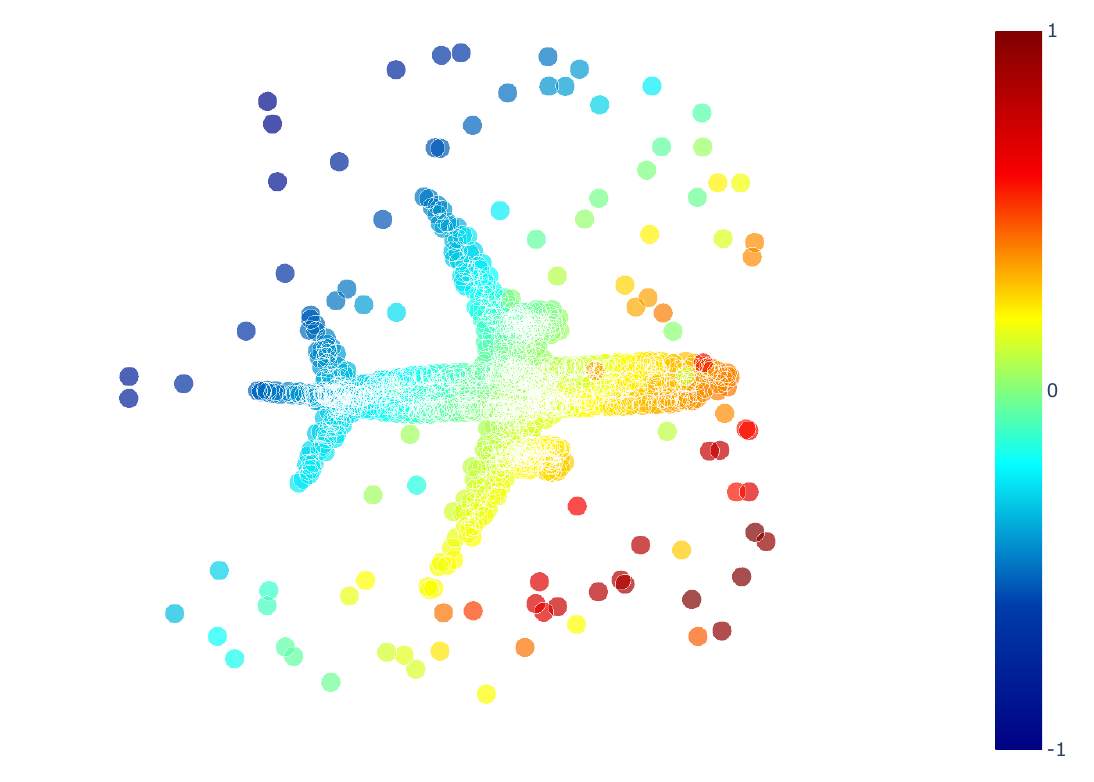} }}%
    \qquad
    \subfloat[\centering Ball $(50\%)$]{{\includegraphics[width=0.3\linewidth]{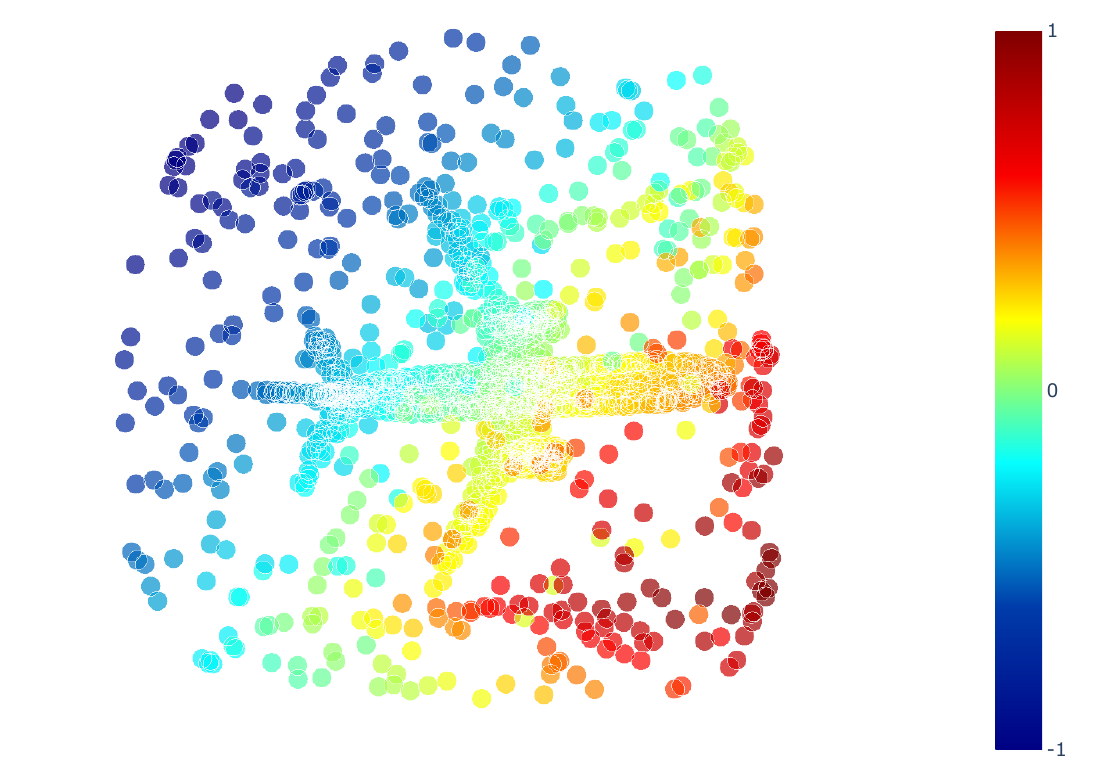} }}%
    \qquad
    \subfloat[\centering Ball $(100\%)$]{{\includegraphics[width=0.3\linewidth]{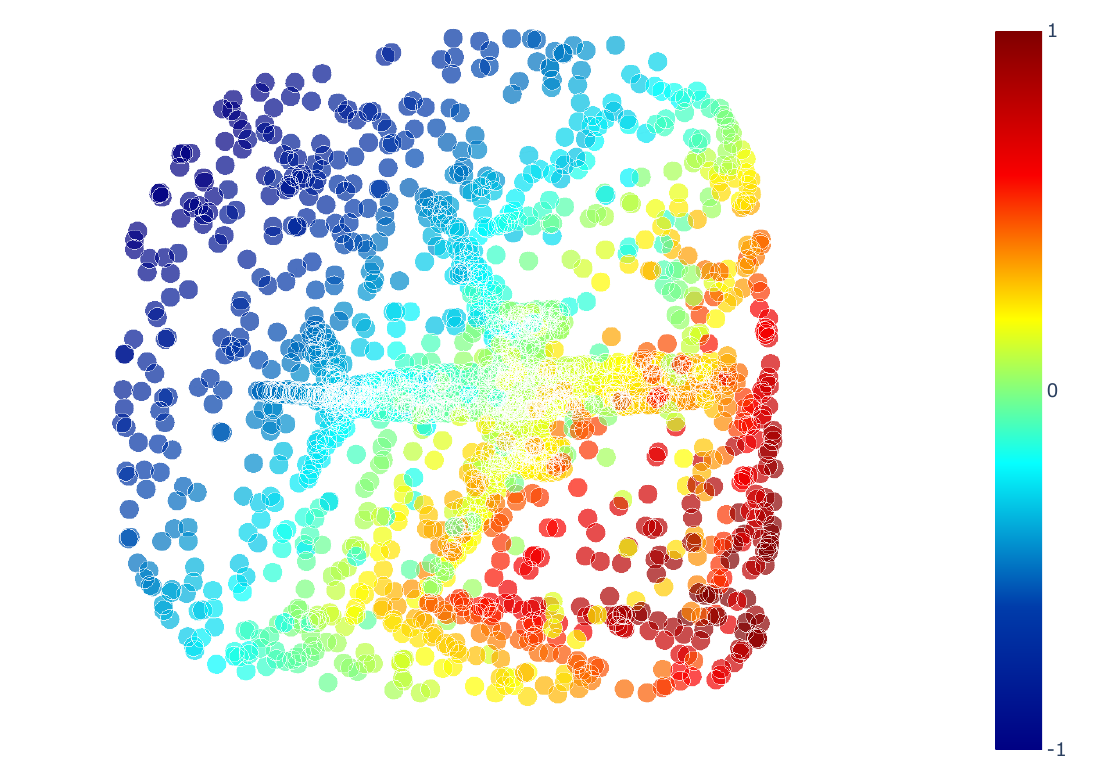} }}%
    \caption{Visualization of noise and outliers. Colors are fake depth for better 3D view.}%
    \label{fig:ball_outliers}%
\end{figure*}

The visualization of our modified ModelNet40-C~\cite{sun2022benchmarking} dataset with noise and outlier corruptions is demonstrated in~\cref{fig:ball_outliers}.

\begin{table*}[t]
\caption[]{Classification results of different models on ModelNet40-C~\cite{sun2022benchmarking} with 3 different data corruptions. 
All results shown here are classification errors. 
The upper tabular between {\color{sol_blue}\textbf{blue bars}} are methods with end-to-end training. 
The lower tabular between {\color{beer_orange}\textbf{orange bars}} are methods with randomly initialized PPEs and trained classifiers.
All the training is on ModelNet40 clean data. }
\centering
\begin{adjustbox}{width=\textwidth}
\begin{tabular}{lcccccccccccccccccc}
\toprule
\multirow{2}{*}{\thead{\normalsize Method}} 
& \multirow{2}{*}{\thead{\normalsize Corruption}} 
& \multicolumn{5}{c}{\thead{\normalsize Density Corruption}}
&& \multicolumn{5}{c}{\thead{\normalsize Noise Corruption}}
&& \multicolumn{5}{c}{\thead{\normalsize Transformation Corruption}} \\
\cmidrule{3-7} \cmidrule{9-13} \cmidrule{15-19}
& (avg.)   & Occlusion & LiDAR & Density Inc. & Density Dec. & Cutout && Uniform & Gaussian & Impulse & Upsampling & Background && Rotation   & Shear  & FFD   & RBF   & Inv. RBF    \\ \midrule
\arrayrulecolor{sol_blue}\toprule[0.3ex]
PointNet~\cite{qi2017pointnet}  & 28.9   & 54.4      & 56.9  & 11.7         & 12.7         & 13.2   && 14.1    & 15.2     & 28.7    & 14.9       & 93.2       && 34.8       & 25.8   & 21.1  & 18.9  & 18.2      \\
PointNet++~\cite{qi2017pointnet++}                                                                & 26.1                                                                     & 56.3      & 71.7  & 17.5         & 11.2         & 11.6   && 24.0    & 19.9     & 43.2    & 21.8       & 22.4       && 27.6       & 15.0   & 15.6  & 17.2  & 16.5      \\
DGCNN~\cite{wang2019dynamic}                                                                & 26.4                                                                     & 60.6      & 82.4  & 14.4         & 18.1         & 15.7   && 16.5    & 18.4     & 25.7    & 16.6       & 51.0       && 20.0       & 12.3   & 13.4  & 15.2  & 14.2      \\
RSCNN~\cite{liu2019relation}                                                                 & 27.8                                                                     & 53.4      & 75.4  & 18.3         & 15.0         & 15.4   && 24.4    & 14.4     & 54.3    & 17.8       & 30.0       && 26.9       & 16.3   & 17.7  & 19.1  & 17.8      \\
PCT~\cite{guo2021pct}                                                                & 24.0                                                                     & 58.2      & 73.4  & 11.8         & 13.8         & 13.3   && 12.0    & 13.7     & 32.1    & 15.5       & 53.7       && 16.4       & 10.8   & 11.6  & 12.2  & 11.6      \\
\arrayrulecolor{sol_blue}\toprule[0.3ex]
\arrayrulecolor{beer_orange}\toprule[0.3ex]
PE-AT (Ours)                                                                 & 26.8                                                                     & 57.8      & 55.6  & 20.1         & 24.3         & 19.0   && 19.8    & 14.9     & 18.1    & 19.3       & 14.9       && 36.3       & 31.5   & 25.3  & 22.9  & 22.5      \\
PE (Ours)                                                                  & 24.1                                                                     & 55.2      & 51.4  & 18.3         & 22.2         & 17.3   && 16.5    & 13.5     & 15.0    & 16.0       & 13.6       && 34.5       & 27.1   & 22.3  & 19.8  & 19.3      \\
\arrayrulecolor{beer_orange}\toprule[0.3ex]
\arrayrulecolor{black}\bottomrule
\end{tabular}
\end{adjustbox}
\label{tab:modelnetc}
\end{table*}

\end{document}